\documentclass{article}

\usepackage{PRIMEarxiv}

\usepackage{graphicx}
\usepackage[many]{tcolorbox}

\usepackage{mathtools,amssymb,latexsym,amsfonts,stmaryrd,bm}
\usepackage{pbox}
\usepackage{xfrac}
\usepackage{booktabs}
\usepackage{xcolor}
\usepackage{threeparttable}
\usepackage{amsthm}
\usepackage{amsmath}
\usepackage{amssymb}
\usepackage{bbding}
\usepackage{hyperref}
\usepackage{multirow}
\usepackage{tikz}
\usepackage{mathpartir}
\usepackage[ruled,linesnumbered]{algorithm2e}
\usepackage{url}

\usepackage{syntax}
\usepackage{framed}
\usepackage[noend]{algpseudocode}
\usepackage{flushend}
\usepackage{listings}
\usepackage{moresize}
\usepackage{wrapfig}
 
\usepackage{seqsplit}
\usepackage{alltt}
\usepackage{xspace}
\usepackage{caption,setspace}
\usepackage{enumitem}
\usepackage{pifont}
\usepackage{makecell}
\usepackage{colortbl}
\usepackage{wrapfig, lipsum}

\usepackage{amsmath, listings, amsthm, amssymb, proof, xspace}

\usepackage{thmtools}

\declaretheoremstyle[spaceabove=\topsep,notefont=\normalfont\itshape]{mystyle}

\newcommand{\revise}[2]{{\color{red}{\ifx&#1&\else- #1\fi}} {\color{ForestGreen}{\ifx&#2&\else+ #2\fi}}}%
\renewcommand{\revise}[2]{#2}%

\usetikzlibrary{arrows,patterns, decorations.pathreplacing}
\newtheorem{definition}{Definition}
\newcommand{\Rom}[1]{(\uppercase\expandafter{\romannumeral #1\relax})}

\newcommand{\F}{Fig.}
\newcommand{\myfig}{Fig.}

\newcommand{\T}{Table}
\renewcommand{\S}{Sec.}
\newcommand{\mysec}{Sec.~}
\newcommand{\A}{Alg.}

\newcommand{\fixme}[1]{\textcolor{black}{#1}}
\newcommand{\ignore}[1]{}

\lstdefinestyle{base}{
  moredelim=**[is][\color{red}]{@}{@},
  escapeinside={<@}{@>}
}

\newcommand{\tool}{\textsc{PDoctor}\xspace}
\newcommand{\name}{\textsc{PDoctor}\xspace}

\newcommand{\parh}[1]{\smallskip\noindent\textbf{#1}}

\usepackage{amssymb}
\usepackage{pifont}

\newcommand\DejaVuttfamily{%
  \fontfamily{DejaVuSansMono-TLF}\selectfont }

\lstdefinestyle{base}{
  moredelim=**[is][\color{red}]{@}{@},
  escapeinside={<@}{@>}
}

\lstdefinelanguage
   [x64]{Assembler}     
   [x86masm]{Assembler} 
   {morekeywords={CDQE,CQO,CMPSQ,CMPXCHG16B,JRCXZ,LODSQ,MOVSXD, %
                  POPFQ,PUSHFQ,SCASQ,STOSQ,IRETQ,RDTSCP,SWAPGS, %
                  rax,rdx,rcx,rbx,rsi,rdi,rsp,rbp, %
                  r8,r8d,r8w,r8b,r9,r9d,r9w,r9b}} 

\usepackage{listings}
\usepackage{fancyvrb}
\usepackage{color}
\definecolor{lightgray}{rgb}{.9,.9,.9}
\definecolor{darkgray}{rgb}{.4,.4,.4}
\definecolor{purple}{rgb}{0.65, 0.12, 0.82}
\definecolor{commentgreen}{RGB}{63,127,95}
\definecolor{redMark}{RGB}{252, 228, 214}
\definecolor{yellowMark}{RGB}{255, 242, 204}
\definecolor{greenMark}{RGB}{226, 239, 218}
\definecolor{blueMark}{RGB}{221, 235, 247}
\definecolor{lightred}{RGB}{255, 130, 130}
\definecolor{myGreen}{RGB}{61, 161, 61}
\definecolor{myBrown}{RGB}{196, 89, 17}
\definecolor{agentBlue}{RGB}{60, 153, 186}
\definecolor{agentGreen}{RGB}{0, 188, 0}
\definecolor{agentBrown}{RGB}{197, 90, 17}

\colorlet{myPurple}{blue!40!red}
\definecolor{myOrange}{RGB}{255,192,0}

\newcommand{\enc}[1]{$\phi^{*}_{\theta}$}
\newcommand{\dec}[1]{$\psi^{*}_{\theta}$}

\lstdefinelanguage{Solidity}{
  keywords={len,delete,int,void,payable, public, event, contract, typeof, new, true, false, catch, function, return, null, catch, switch, var, if, in, while, do, else, case, break,struct,const,socklen_t,sa_familty_t,char,sockaddr},
  keywordstyle=\color{violet}\bfseries,
  ndkeywords={class, export, boolean, throw, implements, import, this},
  ndkeywordstyle=\color{darkgray}\bfseries,
  identifierstyle=\color{black},
  sensitive=false,
  comment=[l]{//},
  escapeinside={(*@}{@*)},          
  morecomment=[s]{/*}{*/},
  commentstyle=\color{commentgreen}\ttfamily,
  stringstyle=\color{red}\ttfamily,
  morestring=[b]',
  morestring=[b]"
}

\newcommand{\rnum}[1]{\uppercase\expandafter{\romannumeral #1\relax}}

\usepackage{xcolor}

\algnewcommand{\LeftComment}[1]{\Statex \(\triangleright\) #1}

\definecolor{pptbrown}{RGB}{132,60,12}
\definecolor{pptgreen}{RGB}{56,87,35}

\let\OLDthebibliography\thebibliography
\renewcommand\thebibliography[1]{
  \OLDthebibliography{#1}
  \setlength{\parskip}{0pt}
  \setlength{\itemsep}{0pt plus 0.1ex}
}

\definecolor{pptgreen}{RGB}{84,130,53}
\definecolor{pptred}{RGB}{176,35,24}
\definecolor{pptblue}{RGB}{194,214,236}

\definecolor{pptgreen1}{RGB}{78,173,91}
\definecolor{pptred1}{RGB}{192,0,0}

\definecolor{pptyellow1}{RGB}{203,195,167}
\definecolor{pptgreen2}{RGB}{184,192,176}

\xspace

\newcommand{\CBrush}{\textcolor[RGB]{84,130,53}{\Checkmark}}
\newcommand{\XBrush}{\textcolor[RGB]{176,35,24}{\XSolidBrush}}

\usepackage{txfonts}

\newif\ifshowcomments
\showcommentstrue

\ifshowcomments
\newcommand{\pc}[1]{\mytodored{[Pingchuan: #1]}}
\else
\newcommand{\pc}[1]{}
\fi
\newcommand{\mytodored}[1]{\textcolor{red}{\ding{46}~{\sf}~#1}}

\newcommand{\kw}[1]{\textcolor{violet}{\texttt{#1}}}
\newcommand{\cf}[1]{\mathbf{#1}}

\title{Testing and Understanding Erroneous Planning in LLM Agents through Synthesized User Inputs}

\author{
  Zhenlan Ji, Daoyuan Wu\thanks{Corresponding authors.}, Pingchuan Ma, Zongjie Li, Shuai Wang\footnotemark[1] \\
  The Hong Kong University of Science and Technology \\
  Hong Kong SAR, China\\
  \texttt{\{zjiae, daoyuan, pmaab, zligo, shuaiw\}@cse.ust.hk} \\
}

\begin{document}
\maketitle

\begin{abstract}
    Agents based on large language models (LLMs) have demonstrated effectiveness
    in solving a wide range of tasks by integrating LLMs with key modules such as
    planning, memory, and tool usage. Increasingly, customers are adopting
    LLM agents across a variety of commercial applications critical to
    reliability, including support for mental well-being, chemical synthesis, and
    software development. Nevertheless, our observations and daily use of LLM
    agents indicate that they are prone to making erroneous plans, especially when
    the tasks are complex and require long-term planning.

    In this paper, we propose \tool, a novel and automated approach to
    testing LLM agents and understanding their erroneous planning. As
    the first work in this direction, we formulate the detection of erroneous
    planning as a constraint satisfiability problem: \textit{an LLM agent's
        plan is considered erroneous if its execution violates the constraints
        derived from the user inputs}. To this end, \tool\ first defines a
    domain-specific language (DSL) for user queries and synthesizes varying
    inputs with the assistance of the Z3 constraint solver. These synthesized
    inputs are natural language paragraphs that specify the requirements for
    completing a series of tasks. Then, \tool\ derives constraints from these
    requirements to form a testing oracle. \tool\ features several design
    considerations, such as mock tool and input mutation, to enhance testing
    effectiveness. Its synthesized inputs can also incorporate advanced
    features like dynamic constraint update
    to better test the LLM
    agent's planning ability.
    We evaluate \tool\ with three mainstream agent frameworks and two powerful
    LLMs (GPT-3.5 and GPT-4). The results show that \tool\ can effectively detect
    diverse errors in agent planning, and provide insights and error
    characteristics that are valuable to both agent developers (for improving LLM
    agents) and users (for using contemporary agents). We conclude by
    discussing potential alternative designs and directions to extend \tool.

\end{abstract}

\section{Introduction}
\label{sec:introduction}

Large language models (LLMs) have become extremely popular due to their remarkable performance across a wide range of tasks~\cite{jiao2023chatgpt, bills2023language, li2022cctest, singhal2023large, CausalityLLMCode23, DeGPT23, GPTScan24}, demonstrating an ability to \textit{understand}, \textit{reason}, and \textit{act} in ways akin to human cognition and reasoning.
With their extensive parameters and training data, these models have shown proficiency in complex pattern recognition in natural language, leading to advancements in logical reasoning~\cite{creswell2022selection, wu2024symbol}, vulnerability detection~\cite{GPTScan24, LLM4Vuln24}, robot planning~\cite{ahn2022can, liu2023llm}, and causal inference~\cite{long2023can, kiciman2023causal}.
This has fueled the development of LLM agents~\cite{yao2022react, AutoGPT, LangChain, shen2024hugginggpt}, which are designed to interact with the external world and make decisions to achieve complex objectives, leveraging LLMs' cognitive capabilities for tasks that require planning, memory, and execution of a sequence of actions.
These agents, integrating key modules alongside LLMs, excel in processing complex user queries, learning from past interactions, and adapting to new tasks.

To date, the industry has been increasingly commercializing LLM agents in various highly profitable and even mission-critical applications~\cite{yao2022react,openaiTools,openaiAssistant,unskript,agentprove,agentfinancial,iAudit24}, such as healthcare, personal assistance, and financial services.
However, despite their great potential and overall enthusiasm, LLM agents often struggle with \textit{erroneous planning}, leading to significant consequences.
For example, an LLM agent used to manage a chemical synthesis process may fail to produce the desired chemical compound if it makes an erroneous plan, and the involved, possibly expensive, chemical reagents are already consumed during the process.
This challenge, highlighted by our observations and experiences, underscores the need for improvements in their design and operation.
Errors in planning, particularly in complex, long-term tasks, can result in the misuse of resources or failure to achieve intended outcomes, underscoring the importance of developing more reliable and effective LLM agents.

In this paper, we propose \name, a novel framework for testing and
understanding erroneous planning in LLM agents. Our approach is fully
automated and can be used to detect erroneous planning in LLM agents using
different core LLMs and following different paradigms (see introduction in
\S~\ref{sec:background}). As the first work in this direction, \tool\
formulates the occurrence of ``erroneous planning'' as a constraint
satisfiability problem: \textit{an LLM agent's plan is erroneous if its
execution violates the constraints derived from the user inputs}. The
constraints can be rigorously checked with moderate cost,
thus providing a reliable way to detect erroneous planning.

\tool\ defines a domain specific language (DSL) that captures the semantics of
user queries and features a synthesis procedure (with the assistance of Z3) that
can generate diverse user requests as the test inputs to the LLM agent. We
provide configurable parameters to control the diversity and complexity of the
user queries and offer a mutation procedure to further transform each generated
user query. Each user query denotes a paragraph that outlines the requirements
for conducting a series of tasks. \tool\ accordingly derives constraints
from these requirements and checks if the LLM agent's plan aligns with the
satisfiability of these constraints;
misalignment flags erroneous planning. We provide a set of design considerations
and optimizations (e.g., mock tools employed by the agent) to deliver
effective testing, and also augment the synthesized inputs with advanced
features like dynamic constraint update
to test the LLM agent's planning capability.

We evaluate \tool\ on three mainstream LLM agent frameworks,
ReAct~\cite{yao2022react}, OpenAI Tools (OT)~\cite{openaiTools}, and OpenAI
Assistant (OA)~\cite{openaiAssistant}. These LLM agents represent different
paradigms and are widely used in various applications. We incorporate these
frameworks with two widely-used LLM models, GPT-3.5 and
GPT-4~\cite{achiam2023gpt}. \tool\ can effectively detect thousands of
erroneous plans across all these settings. We configure \tool\ to depict
the planning capability ``upper bound'' of different LLM agents and also
summarize the detected erroneous plans. These results can provide insights
into the common pitfalls of LLM agents and offer guidance for developers
and users in daily practice. We also discuss the extension of \tool\ and
the potential future work to repair the detected erroneous plans.
In summary, the contributions of this paper are as follows:
\begin{itemize}
    \item We pioneer the effort to test and understand
        erroneous planning in LLM agents, given their increasing
        commercialization in reliability-sensitive fields and the potential
        severe consequences of erroneous planning.
    \item 
        We formulate the problem of detecting erroneous planning as a
        constraint satisfiability problem and present a fully automated
        framework, \tool, that employs input synthesis and constraint
        checking to detect erroneous planning.
    \item We evaluate \tool\ on mainstream LLM agents using different paradigms
        and show that our approach can effectively detect thousands of erroneous
        planning with moderate cost. We also summarize insights from different
        aspects to benefit developers and users.
\end{itemize}

\parh{Tool Availability.} We have made \tool\ available at
\url{https://anonymous.4open.science/r/PDoctor-E872} for review purposes.
We will continue to maintain the tool and add more documentation to assist
users in utilizing it.

\section{Preliminary}
\label{sec:background}

\subsection{Planning Problem}
\label{subsec:planning-problem}

Planning is a fundamental problem-solving task that involves generating a
sequence of actions to achieve a goal~\cite{mccarthy1963situations,
ghallab2004automated}. Extensive efforts have been devoted to this area,
achieving significant progress in various domains, such as
robotics~\cite{nilsson1984shakey, carbonell1991prodigy, joshi2023locally}
and autonomous vehicles~\cite{chen2023interactive, lu2023learning}.
Formally, the planning problem can be defined as follows:

\begin{definition}[Planning Problem]
    Given a set of states $\bm{S}$, a set of actions $\bm{A}$, an and a
    state transition function $f: \bm{S} \times \bm{A} \to \bm{S}$, the
    planning problem $\mathcal{P}$ is defined as a tuple $\langle \bm{S},
    \bm{A}, f, s_0, s_g \rangle$, where $s_0 \in \bm{S}$ is the initial
    state and $s_g \in \bm{S}$ is the goal state. The goal of $\mathcal{P}$
    is to find a sequence of actions $\langle a_1, a_2, \ldots, a_n
    \rangle$ called a plan such that $f(f(\ldots f(s_0, a_1), a_2), \ldots,
    a_n) = s_g$.   
\end{definition}

In the context of LLM agents, the user query can be viewed as a planning
problem $\mathcal{P} = \langle \bm{S}, \bm{A}, f, s_0, s_g \rangle$.
Likewise, the invocation of a specific tool can be viewed as taking an
action $a$, where $a \in \bm{A}$. The goal of LLM agents is to generate a
correct answer, i.e., a sequence of actions $\langle a_1, a_2, \ldots, a_n
\rangle$, to satisfy the constraint set $\bm{C}$ implied by the state
transition function $f$. The initial state $s_0$ represents that the agent
begins to address the user query, and the goal state $s_g$ denotes the
state in which tasks specified in the user query are completed, i.e., all
constraints in $\bm{C}$ are satisfied.

\subsection{An Example of LLM Agents}
\label{subsec:agent-example}

\begin{figure}[!htpb]
    \centering
    \includegraphics[width=1.0\columnwidth]{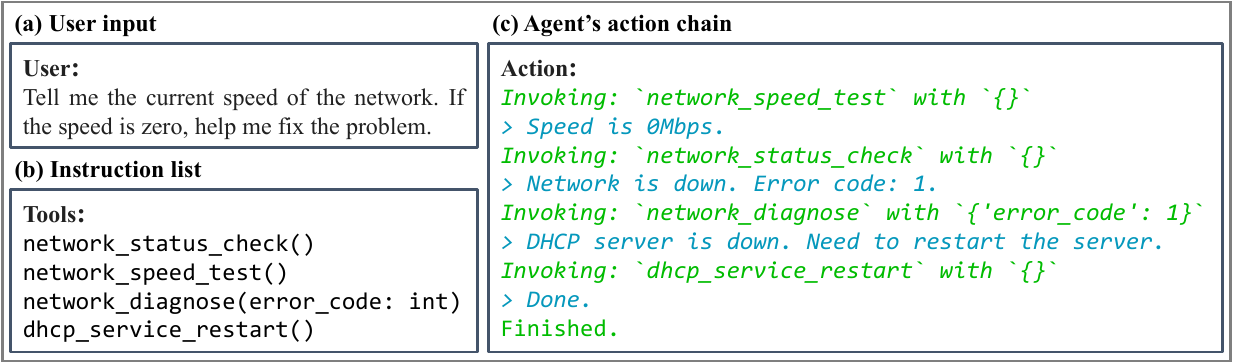}
    \caption{An example illustrating an LLM agent working on a complex user
        query. \textcolor{agentGreen}{Green} and \textcolor{agentBlue}{blue}
        represent the agent's actions and the execution results of invoked tools,
        respectively. }
    \label{fig:agent-exp}
\end{figure}

\F~\ref{fig:agent-exp} presents a typical example of an LLM agent working on a
sample user query (\F~\ref{fig:agent-exp}(a)). The agent first
\textit{understands} the requirements implied by the user query, then plans
(\textit{reasons}) to select the appropriate tools
(\F~\ref{fig:agent-exp}(b)) to invoke, and finally \textit{acts} to
execute each selected tool and collect its response
(\F~\ref{fig:agent-exp}(c)). In particular, after receiving the output
from each invoked tool, the agent repeats the above process until the user query
is fully answered. This series of interactions between the agent and the
environment, as listed in \F~\ref{fig:agent-exp}(c), collectively constitutes a
problem-solving procedure and then determines the final response (omitted
here) to the user query. LLM agents benefit from powerful LLMs that can behave
in a human-like manner to understand, reason, and act simultaneously, and the
synergy of these three capabilities. 

Nevertheless, given the susceptibility of LLMs that has been exposed in prior
research~\cite{LLMLogicalFallacies23,Jailbroken23,HallucinationInevitable24,li2023split},
it is not surprising that LLM agents are substantially prone to errors and
failures. Intuitively, the agent's action chain can be viewed as an ``error
amplifier,'' where a small mistake in the early stage of the action chain could
be continuously amplified and propagated in each subsequent step, leading to
catastrophic failures in the end. This character further exacerbates the
difficulty of providing a correct and stable response to the user query.
Moreover, the interaction between the LLM agent and the environment is often
complex and dynamic, distinct from the static and isolated settings in which
LLMs's abilities have been
tested~\cite{valmeekam2024planbench,li2022cctest,jiao2023chatgpt,LLM4Vuln24}. 

In general, previous studies that endeavor to evaluate LLMs are often limited to
a straightforward linear flow following a ``text in and text out'' paradigm. In
this paradigm, problems are presented in their entirety, and LLMs are expected
to generate a textual response based on the given problem. In contrast, LLM
agent testing necessitates a focus on the interaction and interplay between the
LLM and the environment, where the problems are dynamic and the LLM agent is
required to adjust its plan in response to the environment's feedback. Here, the
feedback refers to the results of the invoked tools, as shown in the blue text
of \F~\ref{fig:agent-exp}(c). In line with these dynamic and complex settings,
we design \tool\ to dynamically update the constraint sets
(see details in \S~\ref{subsec:extended-version}).
This underscores the importance of tool design, in addition to the textual
problems, during test case generation. Consequently, it is crucial to develop a
systematic approach tailored to the unique characteristics of LLM agents to test
and assess their performance in a thorough manner.

\subsection{Architecture of LLM Agents}
\label{subsec:agent-arch}

With a typical example of LLM agents illustrated in \S\ref{subsec:agent-example},
we now delve into the
architecture of an LLM agent to better understand the mechanisms behind the
agent's operation, which is valuable for designing the testing framework.


\begin{figure}[!htpb]
    \centering
    \includegraphics[width=0.80\columnwidth]{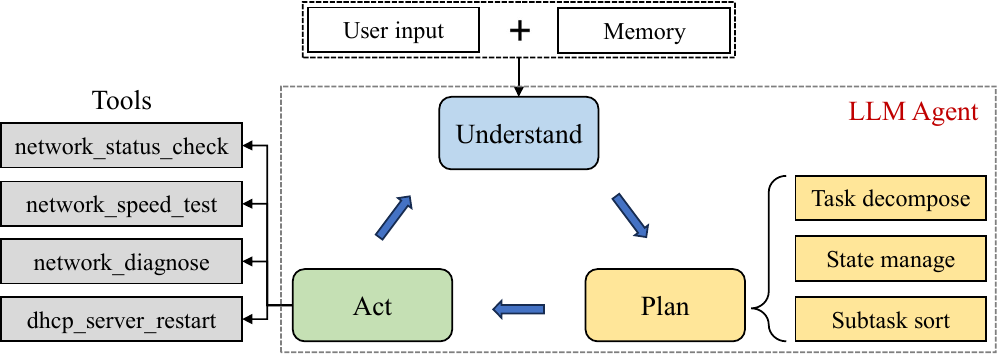}
    \caption{A typical architecture of LLM agents, according
    to~\cite{LLMagentLilian23}. 
    }
    \label{fig:agent-arch}
\end{figure}

\parh{Modules.}~\F~\ref{fig:agent-arch} presents a typical architecture of an
LLM agent. This agent consists of three main components: (1) an
understand module, (2) a plan module, and (3) an act (tool usage)
module. The understand module is tasked with comprehending the user query and
some historical records of previous actions (i.e., in the ``memory'' component)
to extract a concrete task to be accomplished. The plan module is
responsible for: (1) decomposing the extracted task into a sequence of sub-tasks
that can be accomplished by invoking various tools, (2) managing the state and
resources to support an appropriate plan (e.g., retaining the error code until
requested by \texttt{network_diagnose}, as shown in \F~\ref{fig:agent-exp}), and
(3) determining the execution order of the sub-tasks. The act module then
invokes the tools according to the generated plan and collects the results of
the tool invocations.

\parh{LLM and Prompts.}~All the components mentioned above are implemented by an
LLM, which serves as the ``brain'' of the agent. The agent offers prompt
templates composing the query context, how the agent should respond, and the
accessible tools, whereas the users provide the query content. The formed prompt
will be used to guide the LLM to generate the response. 
For example, ReAct~\cite{yao2022react} categorizes the core LLM's response
into three types: \textit{observation}, \textit{thought}, and
\textit{action}, which correspond to the \textit{understand}, \textit{plan}, and
\textit{act} modules in \F~\ref{fig:agent-arch}, respectively.
Additionally, prompt engineering strategies like role-play are often
employed to enhance the agent's performance under specific scenarios, such
as through the \texttt{instruction} parameter in OpenAI's Assistant
API~\cite{openaiAssistant}.


\subsection{Our Testing Focus}
\label{subsec:focus}

As shown in \F~\ref{fig:agent-arch}, different components jointly contribute to
the agent's overall performance. Nevertheless, this paper considers testing the
plan module a top priority. This is because extensive
efforts~\cite{CoTFaithfulness23, CannotSelfCorrect24,
EvaluateInstructionFollowing23, LLM4Vuln24, ReasonWithRules24, CriticBench24}
have been devoted to evaluating the understand module, which rely on LLMs'
basic natural language understanding capability. 
Similarly, the act module has also made promising progress, as evidenced by
OpenAI's recent work in incorporating function calling functionality into GPT
models~\cite{openaiFC}. Moreover, Yue et al.~\cite{huang2023metatool} have
conducted a thorough investigation into LLM agents' tool usage performance and
proposed a benchmark, MetaTool, to evaluate the act module. In contrast, the
plan module, which determines how to concretize the agent's thought
into a sequence of actions that can be executed by the agent, has not been
systematically studied or tested in existing research. To sum up, we have the
following focuses:

\begin{itemize}
    \item \textit{(Isolating the test of the planning module only)} While
    maintaining the complexity of the planning problem, we strive to
    simplify the user queries handled by the understand and act modules
    (see \F~\ref{fig:agent-arch}). By doing this, we can guarantee that any
    detected agent failure can be solely attributed to an error in the
    planning module, thereby achieving a simulated ``isolation'' of the
    planning module. This approach facilitates more precise testing of LLM
    agents' planning ability.

    \item \textit{(Covering both textual queries and underlying tools)}
    Unlike previous
    works~\cite{valmeekam2024planning, valmeekam2024planbench} that tested the
    planning ability of LLMs by generating and evaluating textual plans, the
    planning module in LLM agents is required to decide the order of tool
    invocation, making the testing of the planning module more complex.
    We aim to generate specialized test cases that comprise both textual queries and corresponding tools.

    \item \textit{(Using only valid user queries for testing)}
For individual queries, we focus on generating valid user queries to
test LLM agents. That said, we are \textit{not} using extreme user queries to
stress LLM agents. Based on our observation, extreme queries, such as overly long or
complex ones, as well as those with broken or confusing contents, may hinder
the LLM agent from generating meaningful outputs. This is not
surprising since LLM agents rely on LLMs to generate outputs and are, therefore,
sensitive to the quality of the input prompts.
Sticking to valid user queries can help us better understand the planning
module's performance in a real-world setting.
\end{itemize}

\section{Overview}
\label{sec:overview}

\begin{figure}[!htpb]
    \centering
    \includegraphics[width=0.85\linewidth]{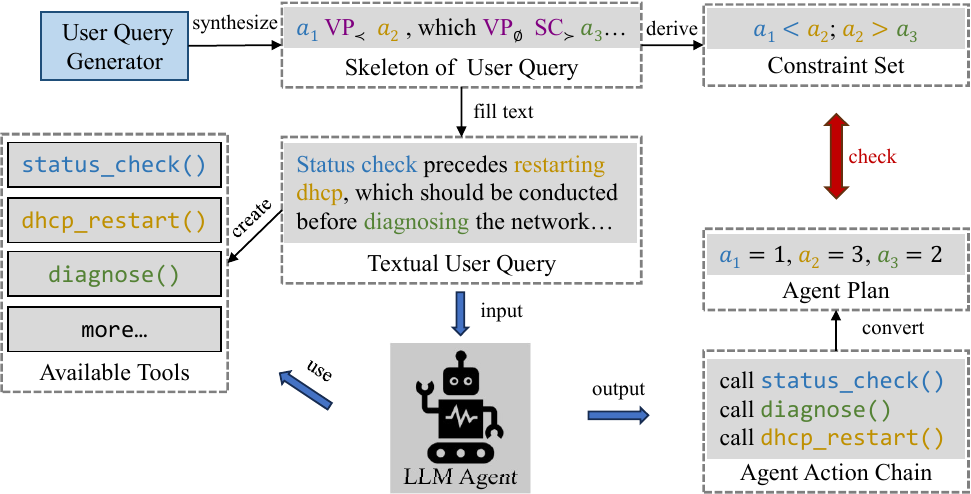}
    \caption{An overall workflow of \tool, which synthesizes textual user queries
    with a list of tools and derives the constraints as the oracle to test
    the LLM agent.}
    \label{fig:workflow}
    \vspace{-5pt}
\end{figure}

With the testing focus described in \mysec\ref{subsec:focus}, we now
introduce \name, a novel system for testing and understanding erroneous
planning in LLM agents. The basic idea of \name is to synthesize
textual user queries, and based on the simultaneously derived formal
constraints, we can identify the erroneous planning.
Based on this idea, we present the overall design of \name as
shown in \myfig~\ref{fig:workflow}.
At a high level, \name generates test cases comprising textual user queries that represent planning problems, along with a series of tools for invocation by the LLM agent.
The testing oracle, automatically derived from the synthesized queries, is used to check the correctness of the LLM agent's planning.
This oracle also facilitates the dissection and characterization of erroneous planning when it occurs.
Specifically, \tool\ has the following key components:

\parh{\ding{172} User Query Generation.}~\tool\ performs a natural language
synthesis process to generate textual user queries, which are considered
as a planning problem $\mathcal{P}$, as formulated in
\S~\ref{subsec:planning-problem}, for the LLM agent. This process comprises
two steps: \textit{skeleton synthesis} and \textit{text
filling}. The former step generates a skeleton of the user query, which is
a high-level, abstract representation of the planning problem. This enables
\tool\ directly to derive the constraints for subsequent testing.
The latter step fills in this skeleton with text to form a complete user
query that describes the requirements for problem-solving. The design of this
synthesis process is motivated by two key considerations: (1) the derived
constraints should be equivalent to the semantics of the generated queries,
since these constraints are used as the
oracle to test the LLM agent; (2) the synthesis process should be flexible
and extensible, allowing users to specify the complexity and scenarios of
the generated queries. For the first consideration, alternative methods 
for generating user queries, such as text mutation and LLM-based
text generation, are further discussed in \S~\ref{sec:discussion}.

\parh{\ding{173} Test Results Check.}~After deriving formal constraints
from the synthesis process, \name utilizes these constraints to detect
erroneous planning of the LLM agent. Specifically, \tool\ maps each action,
which corresponds to subtasks specified by the user query in a one-to-one
manner, to a specific tool invocation and collects the action chain (i.e.,
the planning made by the LLM agent) by recording the history of tool
invocations. The collected planning is then compared against the
constraints to determine whether the planning is correct.

\parh{\ding{174} Error Dissection.}~In contrast to prior studies that
modified inputs based on superficial natural language
properties (e.g., word-level
replacement)~\cite{sun2020automatic,liu2021efficient}, \tool\ synthesize complete textual user queries from scratch,
gaining full control over both the semantics (i.e., the derived constraints) and
the structure (i.e., the skeleton) of the user query. 
This design allows \tool\ to conduct comprehensive mutations and transformations
on the generated query. By maintaining the overall semantic
meaning, \tool\ is capable of altering words, sub-tasks, the context of the
prompt, multiple sentences, and even the entire structure of the
prompt. This capability is crucial for dissecting erroneous planning in 
LLM agents, as it enables \tool\ to pinpoint the root cause of failures
by comparing the planning results of the original and mutated prompts.



\parh{Challenges and Key Solutions.}
To achieve the design above, however, we need to address some unique challenges:
\begin{description}
    \item [C1:] \textit{Designing a mechanism or specification to support
        user query synthesis.} Without a mechanism or specification to
        guide their formulation, the synthesized user queries could become
        highly varied and uncontrollable. Moreover, to easily derive the
        corresponding constraints from the synthesized queries, we also
        need the support of a formal specification. To address this
        problem, we propose a specialized DSL in \mysec\ref{subsec:dsl}.
        With this DSL, \name is able to cover a diverse set of user queries
        with configurable complexity and diversity. Moreover, DSL-based
        skeleton synthesis ensures the full control over the structure and
        semantics of the user queries, enabling \tool\ to exhaustively
        conduct mutation and transformation on a given prompt, thereby
        allowing the dissection of erroneous planning in
        \mysec\ref{subsec:dissection}.

    \item [C2:] \textit{Generating syntactically valid and semantically
        meaningful user queries.} As mentioned in \mysec\ref{subsec:focus}, we
        aim to generate valid user queries to test LLM agents. That said, \tool\
        should focus on generating \textit{syntactically valid} and
        \textit{semantically meaningful} user queries as testing inputs. To
        address this challenge, the user query skeletons are specified using the
        aforementioned DSL, which prevents the generation of broken content; see
        details in \mysec\ref{subsec:synthesis}. Moreover, a constraint solver
        (Z3~\cite{de2008z3} in our case) is used to ensure that the semantics
        underlying the user queries are meaningful, i.e., the constraints
        derived from the generated prompt are always satisfiable, as to be illustrated
        in \mysec\ref{subsec:verify}.

    \item [C3:] \textit{Covering complex planning problems with state management and dynamic problem-solving.}
    While the original design of \name above already covers planning tests for both textual queries and corresponding tools, it has not yet considered complex planning problems involving state management and dynamic problem-solving.
    We address this challenge by introducing time and duration constraints, thereby obtaining an extended testing framework for \name, which will be presented in \mysec\ref{subsec:extended-version}.
\end{description}

\section{Design}
\label{sec:design}



\subsection{Domain-Specific Language}
\label{subsec:dsl}

To support user query synthesis, we have designed a
specialized DSL tailored for LLM agent planning problems.
This DSL is not intended to cover the infinitely vast semantic space of
natural language for describing anything conceivable concept.
Instead, it narrows its focus exclusively to the semantic space required for the planning problems defined in \mysec\ref{subsec:planning-problem}.
Therefore, before delving into the specifics of the DSL, we first present how we simplify the semantic space to be explored, which could also avoid ambiguity due to English not being context-free.

\parh{Semantic Simplification.}~For two given tasks that cannot be executed in
parallel, denoted as task $A$ and task $B$, the planning problem can be
boiled down to identify whether $A \prec B$ (task $A$ comes before task
$B$), or $A \succ B$ (task $A$ comes after task $B$). Accordingly, constraints can
be reduced to a specific description of the execution order of
the tasks. For example, ``\textit{$A$ should be executed before $B$}'' is
a simplified constraint in this context. Moreover, it is straightforward to
generalize this simplification to multiple tasks. Thus, \tool
simplifies prompt synthesis into generating several sentences, each of
which is composed of this kind of simplified constraints.


In addition, each ``task'' in the user query is simplified to take an
action $a$. Here $a \in \mathcal{A}$, and $\mathcal{A}$ is the set of
actions that the LLM agent can perform (see
\S~\ref{subsec:planning-problem}). Take \F~\ref{fig:workflow} as an
example, conducting network status check is a task and invoking
\texttt{network\_status\_check()} is the action. Without necessitating
additional decomposition, every task can be completed directly through one
single action. This simplification is based on the observation that task
decomposition is mainly determined by the context of the task, rather than
the performance of the LLM agent itself. For example, LLM agents that are
familiar with the network would be able to correctly decompose the task of
``fix disconnected network'' into a series of sub-tasks like ``check
network status'' and ``network diagnosis'', while others that are designed
for addressing book management would not. Nonetheless, this task decomposition
failure can be easily mitigated by providing the LLM agent with instructive
knowledge of network repair.  Due to the potential unanticipated impact
that task decomposition could have on the performance of the LLM agent,
\tool\ avoids generating complex tasks that require decomposition.

\begin{figure}[t]
    \centering
    \hrule
    \smallskip
    \begin{minipage}[t]{0.90\linewidth}
        \textbf{Syntax}
        \begin{align*}
            \langle \mathcal{P} \in \text{Planning Problem} \rangle \Coloneqq{} & S^{+}                                                                             \\
            \langle S \in \text{Sentence} \rangle \Coloneqq{}                   & s \mid S\; j\; s                                                                  \\
            \langle s \in \text{Sub-sentence} \rangle \Coloneqq{}               & c_{i} \mid m                                                                      \\
            \langle c_{i} \in \text{Independent Clause} \rangle \Coloneqq{}     & s\; (\kw{VP}_{\prec} \mid \kw{VP}_{\succ}) \; o                                   \\
            {}\mid{}                                                            & s\; \kw{VP}_{\emptyset}\; (\kw{P}_{\prec} \mid \kw{P}_{\succ})\; o                \\
            {}\mid{}                                                            & (\kw{P}_{\prec} \mid \kw{P}_{\succ})\; o\; \text{``,''}\; s\; \kw{VP}_{\emptyset} \\
            \langle m \in \text{Multi-clauses} \rangle \Coloneqq{}              & c_{d}\; (\kw{SC}_{\prec} \mid \kw{SC}_{\succ})\; c'_{d}                           \\
            {}\mid{}                                                            & (\kw{SC}_{\prec} \mid \kw{SC}_{\succ})\; c'_{d}\; \text{``,''}\; c_{d}            \\
            \langle c_{d} \in \text{Dependent Clause} \rangle \Coloneqq{}       & s\; \kw{VP}_{\emptyset}                                                           \\
            \langle c_{r} \in \text{Relative Clause} \rangle \Coloneqq{}        & \text{``which''}\; (\kw{VP}_{\prec} \mid \kw{VP}_{\succ})\; o                     \\
            {}\mid{}                                                            & \text{``which''}\; \kw{VP}_{\emptyset}\; (\kw{P}_{\prec} \mid \kw{P}_{\succ})\; o \\
            \langle s \in \text{Subject} \rangle \Coloneqq{}                    & a^{+}                                                                             \\
            {}\mid{}                                                            & a^{+}\; \text{``,''}\; c_{r}\; \text{``,''}                                       \\
            \langle o \in \text{Object} \rangle \Coloneqq{}                     & a^{+}                                                                             \\
            {}\mid{}                                                            & a^{+}\; \text{``,''}\; c_{r}\; \text{``,''}                                       \\
            \langle j \in \text{Conjunction} \rangle \Coloneqq{}                & \text{``;''}                                                                      \\
            {}\mid{}                                                            & \text{``,''}\; (\text{``and''} \mid \text{``but''} \mid \text{``yet''})           \\
            {}\mid{}                                                            & \text{``,''}\; (\text{``while''} \mid \text{``whereas''})                         \\
            \langle a \in \text{Action} \rangle \Coloneqq{}                     & \text{the set of actions, } \mathcal{A}                                           \\
        \end{align*}
    \end{minipage}%
    \vspace{-5pt}
    \hrule
    \captionsetup{skip=5pt}
    \vspace{5pt}
    \caption{The syntax of our DSL, specifically designed for synthesizing user queries and deriving their constraints.}
    \label{fig:dsl}
\end{figure}

\parh{DSL Specifics.}~\F~\ref{fig:dsl}
presents the syntax of this domain-specific language.
Overall, each user query is represented as a paragraph $\mathcal{P}$. Every
task that the LLM agent is expected to perform is detailed in this
paragraph; each task requires the LLM agent to take a specific action $a$.
Paragraphs are composed of one or more sentences $S$. And each sentence $S$
is a sequence of sub-sentences $s$ separated by a clause conjunction $j$. A
sub-sentence can be regarded as an independent unit of semantic meaning,
containing one or multiple constraints to be obeyed by the LLM agent. In
particular, a sub-sentence $s$, no matter if it is an independent clause
$c_{i}$, or a multi-clause $m$, comprises a subject $s$, an object $o$, and
one or more keywords in \kw{purple} that directly indicate a sequential
relationship. These keywords are denoted in a special way, where the
capital letter stands for their part-of-speech (POS) tag, and the subscript
denotes the sequential relationship. Specifically, $\kw{VP}$ denotes a verb
phrase, $\kw{P}$ denotes a preposition, and $\kw{SC}$ denotes a sequential
conjunction. Here, ``sequential conjunction'' is a special type of
conjunction that indicates the temporal relationship between two dependent
clauses. For the subscript, we use $\prec$ to stand for the case where the
events in the subject clause happen before the events in the object clause,
while $\succ$ denotes the opposite. $\emptyset$ is an exceptional case, as
it is merely used to decorate the verb phrase $\kw{VP}$, indicating that
the verb phrase is not associated with any temporal relationship, like
``happen'' or ``occur''.

\subsection{User Query Synthesis}
\label{subsec:synthesis}

The DSL designed in \mysec\ref{subsec:dsl} moderates natural language complexity
while preserving the expressiveness and variety of the user queries it generates. In general,
synthesizing a user query is to gradually expand an abstract syntax tree
(AST) where each node in the AST is randomly selected from a set of valid
nodes according to the grammar of the DSL. That said, the synthesis process
is conducted in a top-down manner, with the root node being the paragraph
$P$, and the leaf nodes being a sequence of non-terminal symbols in the
DSL. Without accounting for the expansion of $a^{+}$, the DSL offers a total
of 340 possible expanding options, ensuring the diversity of the
synthesized user queries.

Based on this carefully designed DSL, the synthesis of user queries mainly comprises three steps: synthesizing the skeleton, translating the skeleton into NL (natural language) prompts, and simultaneously deriving the constraints from the synthesized skeleton, as illustrated in \myfig~\ref{fig:workflow}.

\begin{algorithm}[!htbp]
    \footnotesize
    \caption{Skeleton Synthesis.}
    \label{alg:synthesis}
    \KwIn{Action Set $\bm{A} = \{a_1, a_2, ..., a_n\}$, Max Sentence Number
        $N$, Max Iteration $K$}
    \KwOut{Generated Paragraph $P$, Constraint Set $\bm{C}$}

    $P \gets \emptyset$\\
    $\bm{C} \gets initConstraints(\bm{A})$ \tcp{Initialize constraints.}
    \For{$n \in \{1, ..., N\}$}{
        \tcc{Iteratively generate sentences to form a paragraph.}
        $\texttt{iter} \gets 0$\\
        \While{$\texttt{iter} < K$}{
            $s \gets genSentence(\bm{A}, \bm{C})$\\
            $\bm{C}' \gets \bm{C} \cup getConstraints(s)$\\
            \If{\normalfont $SATSolver(\bm{C}') == \text{SAT}$}{
                \tcc{Verify the satisfiability of the constraints after adding the new sentence.}
                $P \gets P \cup s$\\
                $\bm{C} \gets \bm{C}'$\\
                \textbf{break}
            }
            $\texttt{iter} \gets \texttt{iter} + 1$\\
        }
        \If{$\texttt{iter} \geq K$}{
            \textbf{break}
        }
    }
    \Return{$P, C$}
\end{algorithm}

\parh{Skeleton Synthesis.}~\A~\ref{alg:synthesis} presents the algorithmic
procedure for prompt skeleton synthesis. The algorithm takes the action set
$\bm{A}$, the maximum number of sentences $N$, and the maximum iteration
$K$ as inputs, and outputs the generated paragraph $P$ and its
corresponding constraint set $\bm{C}$. In general, for each sentence
generation, the algorithm iteratively generates a sentence by randomly
expanding the sentence $S$ according to the DSL and then substituting
non-terminals with natural language words (line 6). For action symbols in
the skeleton, they are substituted with actions from $\bm{A}$. Constraints
of this sentence are then derived and used to extend the constraint set
$\bm{C}$ (line 7). If the extended constraint set $\bm{C}'$ is satisfiable,
the sentence will be added to paragraph $P$, and the constraint set
$\bm{C}$ will get updated (lines 8--11). Otherwise, the algorithm will try
to generate another sentence by iterating the process until the maximum
iteration $K$ is reached, at which point the generation process will be
terminated since it indicates that finding a satisfiable sentence is
infeasible (lines 12--17). It is worth noting that this algorithm ensures
the generated user query must be satisfiable, as the constraints are checked
after each sentence generation. This way, the synthesized user query can
effectively be used for testing the LLM agent's planning ability.


\parh{Text Filling.}~To ``translate'' the skeleton into natural language
user queries, \tool\ fills text into the slots marked by the terminal
symbols (e.g., various \kw{keywords} and actions) in the skeleton. Except
for action symbols, there are seven alternative natural language words on
average for each terminal symbol. For example, ``happen'', ``occur'', ``be
executed'', etc., for the verb phrase $\kw{VP}_{\emptyset}$, and ``after'',
``behind'', ``later than'', etc., for the preposition $\kw{P}_{\succ}$. For
the action symbols, we substitute them with the daily activities of various
jobs. More specifically, we first instruct the LLM to provide a list of
common jobs (e.g., ``teacher'', ``software developer'', etc.) by querying
the LLM with the prompt ``\textit{Please provide a list of 50 typical jobs.
    Ensure that these 50 positions span a variety of industries.}''. For each
provided job, we further query the LLM with the prompt ``\textit{Please
    list 20 activities in noun phrase format that a [role] may need to do in a
    day.}'', where ``[role]'' is replaced with the job name. Action symbols in
the same paragraph are substituted with daily activities from the same job,
and the job becomes the \textit{Topic} of the user query. By changing the
topic, \tool\ is capable of smoothly altering the context of the user
queries, which is beneficial for the error dissection in
\mysec\ref{subsec:dissection}. The design of text filling facilitates the
diversity and realism of the synthesized user queries, in contrast to the
previous works that are limited to a few fixed, artificial scenarios or
templates~\cite{valmeekam2024planbench, shridhar2020alfworld}.



\parh{Deriving Constraints.}
Simultaneously, \name derives the constraints from each synthesized user query.
Specifically, each action $a$
is represented as an integer variable $\cf{a}$, denoting the execution order of
the action. Then, the subsequent relationship between two actions, denoted as
$a_1$ and $a_2$, can be expressed in a comparative form, like $\cf{a_1} <
    \cf{a_2}$ or $\cf{a_1} > \cf{a_2}$. The former indicates that action $a_1$
should be executed before action $a_2$, while the latter denotes the opposite.
Therefore, considering a sub-sentence $s$, ``\textit{network diagnosis comes
    after network status check}'' with the skeleton $a_1\, \kw{VP}_{\succ}\, a_2$,
the constraint derived from this sub-sentence is $\cf{a_1} > \cf{a_2}$.

\subsection{Test Results Check}
\label{subsec:verify}

Once we feed the synthesized user query to the LLM agent, it will generate a
textual response that describes the planning of the tasks. We can check
the correctness of the planning by comparing the action sequence extracted
from the response to the constraints derived from the user query.
It is worth noting that some may argue for employing
metrics like the BLEU score~\cite{papineni2002bleu} and
BertScore~\cite{zhang2019bertscore} to assess the similarity between the
generated response and the ground truth. However, this is less feasible in
our context because (i) these metrics may be biased or expensive to
compute, and (ii)
the effectiveness of planning testing should be assessed based on whether the action sequence satisfies the constraints, rather than on the similarity between the generated response and the ground truth.
Our approach of comparing the action sequence to the constraints, instead,
is a more direct and effective way to validate the LLM agent's planning.

A new challenge arises, however, as extracting the action sequence from the
LLM agent's textual response is non-trivial, requiring a precise
understanding of the LLM response meaning. Existing work typically
instructs the LLM to respond in a structural format through the adoption of
few-shot prompts~\cite{liu2023llm, valmeekam2024planbench}. We argue,
nevertheless, that this practice is still not ideal, as LLMs may fail to
comply with the format requirements. For example, the LLM may generate a
response that is not structured as expected, or the response may contain
irrelevant information that complicates the extraction process.

\begin{figure}[t]
    \centering
    \includegraphics[width=0.6\columnwidth]{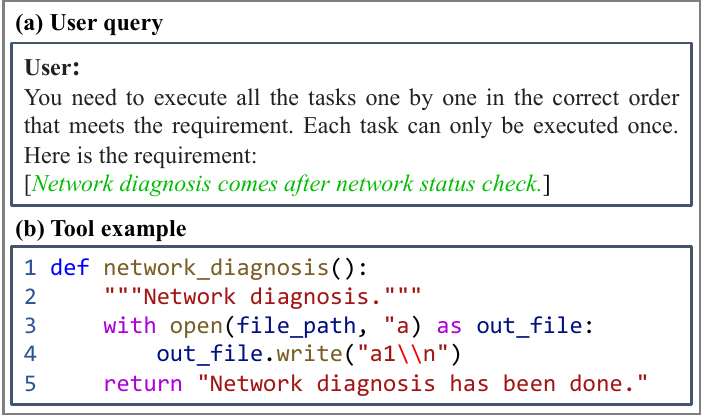}
    \caption{An example illustrating our ``mock test'' approach to
        extract the action sequence of LLM agents.
        \textcolor{agentGreen}{\textit{Green italic}} denotes the synthesized
        part, which describes the constraints for conducting a series of
        tasks.}
    \label{fig:basic-exp}
\end{figure}

\parh{Mock Tool Design.}~\tool\ conducts a ``mock test'' approach to extracting
the action sequence of LLM agents.
Specifically, for each action $a$ in the synthesized user query, \tool\
creates a ``simulated'' tool, with the tool's name and description
designed to be consistent with the textual content of the action (e.g.,
``network diagnosis'' in \F~\ref{fig:basic-exp}). The internal logic of the
tool is designated as a simple file-write module that directly writes the
ID of the corresponding action $a$ into a log file. This design is based on
the fact that LLM agents treat tools in a black-box manner: they select
tools based on the tool's name and description, rather than the internal
implementation. Therefore, the literal task described by $a$ is not
conducted in the environment, and the tool merely returns a string that
deceives the LLM agent into believing that the action has been completed.
\F~\ref{fig:basic-exp} illustrates an example of this design. For the
synthesized user query presented in \F~\ref{fig:basic-exp}(a), \tool\
creates a set of tools corresponding to the actions in the user query,
where \F~\ref{fig:basic-exp}(b) depicts one of the tools. The
\texttt{network\_diagnosis} does not perform an actual network diagnosis in
the environment. Instead, it merely writes the tool's ID, ``a1'', into the
log file and deceives the LLM agent into believing that the network has
been diagnosed.

\parh{Result Checking.}~After the agent finishes processing the synthesized user
query, the log file is read to collect the action sequence. Take the case
in \F~\ref{fig:workflow} as an example, the action sequence is ``[$a_1$,
$a_2$, $a_3$]''. Then, for each action $a$ in the action sequence, the
corresponding variable $\cf{a}$ is assigned with the execution order of the
action, i.e., $\cf{a_1} = 1$, $\cf{a_2} = 3$, and $\cf{a_3} = 2$.
Afterward,
a plan is identified as erroneous if there exists any constraint
unsatisfied by the assigned values of the variables.
Still referring to the example in \F~\ref{fig:workflow}, the assigned values of
variables $\cf{a_1}$, $\cf{a_2}$, and $\cf{a_3}$ are checked against the logic
and ($\land$) of the constraint set $\{\cf{a_1} < \cf{a_2}$, $\cf{a_2} >
    \cf{a_3}\}$. Since none of the constraints (and their $\land$) is unsatisfied,
this planning is identified as correct. Overall, we view the integration of this
check design with the mock tools as a novel approach to rigorously examine the
planning of LLM agents while avoiding the complexity of understanding LLMs'
textual response.

\subsection{Extended Testing Framework}
\label{subsec:extended-version}

\begin{figure}[t]
    \centering
    \hrule
    \smallskip

    \begin{minipage}[t]{0.99\linewidth}
        \textbf{Syntax}
        \begin{align*}
            \langle o \in \text{Object} \rangle \Coloneqq{} & t                                          \\
            \langle t \in \text{Time} \rangle \Coloneqq{}   & \text{the set of time point, } \mathcal{T} \\
        \end{align*}
    \end{minipage}%
    \vspace{-5pt}
    \hrule
    \vspace{5pt}
    \captionsetup{skip=5pt}
    \caption{DSL extension for covering time and duration constraints.}
    \label{fig:ext-dsl}
\end{figure}

\begin{figure}[t]
    \centering
    \includegraphics[width=1.0\columnwidth]{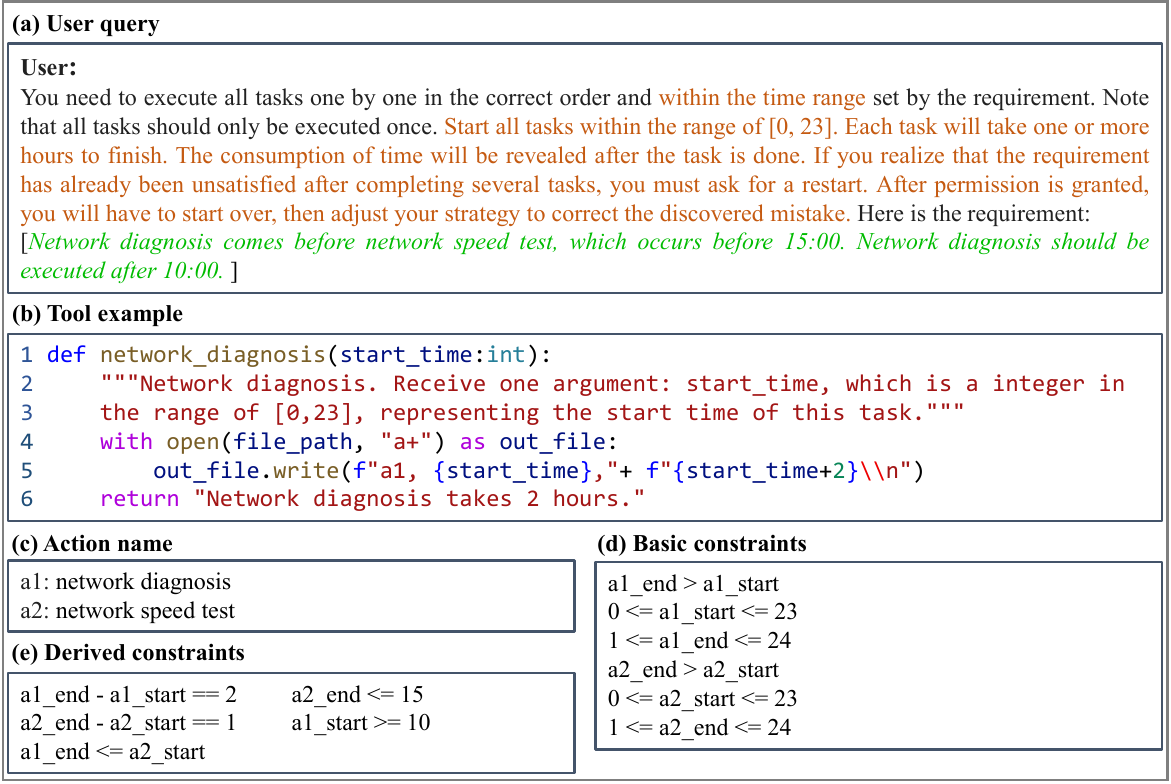}
    \caption{An example illustrating the extended version of the generated
        test case, including the user query and corresponding tools.
        Similar to \myfig~\ref{fig:basic-exp},
        \textcolor{agentGreen}{\textit{green italic}} denotes the
        synthesized part. Newly added instructions are highlighted in
        \textcolor{agentBrown}{brown}. Sub-figure (c) shows the mapping
        between the action ID and the action name. (d) and (e) present the
        basic constraints and the additional constraints derived from the
        user query, respectively.}
    \label{fig:extended-exp}
\end{figure}

Although the aforementioned test framework is effective in evaluating the
planning ability of LLM agents, real-world scenarios would be more complex
and challenging. Referring to the example in \F~\ref{fig:agent-exp}, the
agent is expected to dynamically adjust its planning based on the execution
results of the tools. Moreover, some tools may require several specific
inputs, like \texttt{error\_code} in this example, which requires the LLM
agent to retain the state information across different tasks. Motivated by
these observations, we extend the test framework by introducing time and
tool duration concepts into the agent planning. For one thing, time
constraints are widely used in real-world planning, such as daily
schedules. For another, the introduction of time allows \tool\ to simulate
the dynamic planning over global resources, which is a representative
complex planning problem in real-world scenarios. In particular, the
synthesized user query is extended to include new constraints that regulate
the start or end time of given tasks as illustrated in
\F~\ref{fig:ext-dsl}. Accordingly, the mock tools are modified to
require the start time point as the input parameter and return the duration
of their execution. \F~\ref{fig:extended-exp} illustrates an example of the
extended version of the generated test case.


From this figure, an obvious change is the template of the user query.
Several new instructions are added, to require the LLM agent to consider
the time constraints. Besides, the change in the return value of the
mock tool, which is presented in \F~\ref{fig:extended-exp}(b), requires
the agent to dynamically adjust its planning. Due to the absence of knowledge about the
tool execution time, the agent may encounter a situation where the
planning is negated by the execution result of tools. Taking the case in
\F~\ref{fig:extended-exp} as an example, if the agent starts the execution
of the \texttt{network\_diagnosis} tool at 14:00, this planning would be
negated because \texttt{network\_diagnosis} takes 2 hours and finishes at
16:00, while the subsequent action \texttt{network\_speed\_test} is
required to be executed before 15:00. In this case, the agent is instructed
to conduct an early halt and re-plan as shown in
\F~\ref{fig:extended-exp}(a).

Besides, the constraints derivation also changes. Now each action $a$ is
associated with two integer variables $\cf{a}\_{\text{start}}$ and
$\cf{a}\_{\text{end}}$, denoting the start and end time of the action,
respectively. A new kind of constraint, duration constraint, is derived
from the mock tools' return value, regulating the duration of the
associated action. \F~\ref{fig:extended-exp}(d) shows the basic constraints
that are implied by the test environment and the instructions in the user
query. \F~\ref{fig:extended-exp}(e) presents the additional constraints
derived from the synthesized part that is highlighted in
\textcolor{agentGreen}{\textit{green italic}}.



\begin{algorithm}[!htbp]
    \footnotesize
    \caption{Error Dissection.}
    \label{alg:error-dissection}
    \KwIn{User Query $Q$ that triggers the erroneous planning, Tool Set $\bm{T}$, Constraint Set: $\bm{C}$, LLM Agent $\textsc{A}$, Max Iterations $K$}
    \KwOut{Error Cause $E$}


    \For{$n \in \{1,2,3\}$}{
        \tcc{Identify whether the error emerges probabilistically.}
        $\texttt{plan} \gets \textsc{A}(Q, \bm{T})$\\
        \If{\normalfont $CheckPlan(\texttt{plan}, \bm{C}) == \text{True}$}{
            \Return{\normalfont ``Probability''}
        }
    }

    \For{$k \in \{1, ..., K\}$}{
        \tcc{Mutate the user query by substituting the textual content of terminals in it.}
        $Q',\bm{T}' \gets TerminalSubstitute(Q,\bm{T})$\\
        $\texttt{plan} \gets \textsc{A}(Q', \bm{T}')$\\
        \If{\normalfont $CheckPlan(\texttt{plan}, \bm{C}) == \text{True}$}{
            \Return{\normalfont ``Terminal''}
        }
    }
    \For{$k \in \{1, ..., K\}$}{
        \tcc{Mutate the user query by changing its topic.}
        $Q', \bm{T}' \gets TopicChange(Q, \bm{T})$\\
        $\texttt{plan} \gets \textsc{A}(Q', \bm{T}')$\\
        \If{\normalfont $CheckPlan(\texttt{plan}, \bm{C}) == \text{True}$}{
            \Return{\normalfont ``Topic''}
        }
    }
    \For{$k \in \{1, ..., K\}$}{
        \tcc{Synthesize a new test case whose constraints are equivalent to the original one.}
        $Q', \bm{T}' \gets QuerySynthesize(\bm{C})$\\
        $\texttt{plan} \gets \textsc{A}(Q', \bm{T}')$\\
        \If{\normalfont $CheckPlan(\texttt{plan}, \bm{C}) == \text{True}$}{
            \Return{\normalfont ``Structure''}
        }
        \Else{
            \Return{\normalfont ``Constraint''}
        }
    }



\end{algorithm}

\subsection{Error Dissection}
\label{subsec:dissection}

\tool\ also features an error dissection component designed to investigate the
causes of erroneous planning in LLM agents. Recall that our proposed query
synthesis technique facilitates a thorough understanding of query semantics and
the constraints that the LLM agent should satisfy. Given an error-triggering
query, \tool\ can deliberately mutate components in this query to dissect the
triggered error.

\A~\ref{alg:error-dissection} presents the error dissection algorithm. In
general, this algorithm first checks whether the error emerges probabilistically
by running the LLM agent multiple times with the same user query. It then
proceeds to mutate the user query from a low-level word substitution to
high-level modification to identify the causes of the error. Specifically, the
algorithm employs three mutation strategies: \textit{TerminalSubstitute},
\textit{TopicChange}, and \textit{QuerySynthesize}. Correspondingly, there are
five possible error causes: \textit{Probability}, \textit{Terminal},
\textit{Topic}, \textit{Structure}, and \textit{Constraint}. If the error
happens probabilistically, the error cause is \textit{Probability} (lines 1--6).
If the error is caused by specific words and can be fixed by substituting these
words, the error cause is \textit{Terminal} (lines 7--13). If the error can only
be rectified by changing the topic of the user query, the error cause is
\textit{Topic} (lines 14--20). Furthermore, \tool\ will try to synthesize a new
user query whose constraints are equivalent to the original one. If the error
does not occur with the new query, the cause is \textit{Structure}; otherwise,
it is \textit{Constraint} (lines 21--30). \textit{Constraint} means that \tool\
reveals a special set of constraints that the LLM agent is more prone to make
mistakes with. Note that we do not consider dissecting multiple errors over a
query, as this may complicate the whole process and make it hard to pinpoint the
genuine cause of the error. For instance, if the error behaves
probabilistically, it becomes less meaningful to dissect the error further.
Similarly, if we have already identified specific words that cause the error
(i.e., \textit{Terminal}), then it should not belong to more holistic causes
like \textit{Topic} or \textit{Structure}.

\section{Implementation and Experiment Setup}
\label{sec:setup}

\tool\ is implemented in Python3 with about 2,600 lines of code. We
integrate \tool\ with Z3~\cite{de2008z3}, a popular constraint solver, for
user query synthesis and mutation. In addition, all LLM agents are implemented
using LangChain~\cite{LangChain}, a prevalent Python-based framework for
developing LLM agents.

\parh{Models.}~Two widely-used LLM models, GPT-3.5 and GPT-4, are employed in
our evaluation for two main reasons. First, acting as the agent core to
conduct correct planning is challenging, thereby necessitating the need for
powerful LLMs like the ones from OpenAI. Second, OpenAI's models are the
mainstream choices among the LLM application community, with the majority of
popular libraries (e.g., LangChain) using them as the default. In this paper,
all parameters of the LLM models are set to their default values except for
\texttt{temperature}, which is set to $0$ to mitigate the non-determinism of LLM
responses (OpenAI Assistant excludes this parameter because it does not support
it). The versions of both models employed in this study are set to 1106, namely \texttt{gpt-3.5-turbo-1106} and \texttt{gpt-4-1106-preview}.

\parh{LLM Agent Frameworks.}~We evaluate \tool\ on three mainstream LLM agent
frameworks, including ReAct~\cite{yao2022react}, OpenAI Tools
(OT)~\cite{openaiTools}, and OpenAI Assistant (OA)~\cite{openaiAssistant}. These
LLM agents are selected to cover different design choices and paradigms,
including different interaction modes (how the LLM agent interacts with the core
LLM or the integrated tools), prompt template designs, and in-depth
customizations (e.g., LLM sampling parameter tuning, memory management, etc.).

\T~\ref{tab:agents} lists the details of the evaluated LLM agent frameworks.
ReAct is the most compatible agent framework, as it supports any LLM that can be
invoked via a \texttt{completion}. In contrast, OT and OA require the core LLM to
support chat-based interaction and be tuned with function calling capabilities~\cite{openaiFC}.
Both ReAct and OT are built with an emphasis on
flexibility, allowing users to customize the LLM agent in-depth and \fixme{use
prompt templates to guide the user queries.}
In contrast, OA is designed to be \fixme{more ``fool-proof''}, with limited
customization options. This simple design choice has garnered considerable
attention for OA among the general public and may signify a new trend in the LLM
agent sector. To the best of our knowledge, the aforementioned LLM agents are the most
representative ones in the current LLM agent landscape. It is noteworthy,
nevertheless, that \tool\ is not limited to these LLM agent frameworks and can
be applied to other LLM agents as well.

\begin{table}[t]
    \centering
    \caption{Details of the evaluated LLM agents.}
    \label{tab:agents}
    \resizebox{0.9\linewidth}{!}{
        \begin{tabular}{l|c|c|c|c}
            \toprule
            \textbf{Tool}             & \textbf{LLM Interaction} & \textbf{Tool Interaction} & \textbf{Prompt Template?} & \textbf{In-depth Customization?} \\
            \midrule
            ReAct~\cite{yao2022react} & Completion               & Few-Shot Prompting        & \CBrush                   & \CBrush                          \\
            OT~\cite{openaiTools}     & Chat                     & Function Calling          & \CBrush                   & \CBrush                          \\
            OA~\cite{openaiAssistant} & Chat                     & Function Calling          & \XBrush                   & \XBrush                          \\
            \bottomrule
        \end{tabular}
    }
\end{table}

\section{Evaluation}
\label{sec:evaluation}

In this section, We aim to answer the following research questions (RQs):
\begin{description}
    \item [RQ1:] How effective is \tool\ in detecting erroneous planning in LLM agents?

    \item [RQ2:] How well do LLM agents perform planning and what kinds of planning errors do they make?

    \item [RQ3:] How do LLM agents perform when encountering complex planning problems?
\end{description}



\subsection{RQ1: Assessing \name's Effectiveness in Detecting Erroneous Planning}
\label{sec:rq1}

We first assess the effectiveness of \tool\ in detecting erroneous
planning in LLM agents. For each type of LLM agent (six in total, two models
for three types of LLM agent frameworks as described in
\S~\ref{sec:setup}), we employ \tool\ to randomly generate test cases and
test the agent's planning performance within a specified time constraint.
Since token usage increases exponentially with agent iteration,
rendering the test of LLM agents costly, we set the time limit to
60 minutes for each type of LLM agent. For each generated test case, we
create a new LLM agent instance and instruct it to process the synthesized
user query with the provided tools, with the timeout configured to 180
seconds. The maximum iteration limit is set to 50, consistent with that in Yao et
al.'s work~\cite{yao2022react}.

Moreover, the difficulty level of the planning problem is a key factor that
affects the planning performance of LLM agents. In particular, the LLM
agent is a difficulty-sensitive system expected to perform better on easier 
problems and worse on harder problems. The rationale behind this
is that the ultimate goal of LLM agents is to generate human-like text by
imitating the logic underlying human behaviors, and human beings are known
to be difficulty-sensitive when solving problems. \tool\ takes the action
number, $|\mathcal{A}|$, as the key parameter to control the difficulty of
the generated planning problem. Since the agent is required to invoke every
tool once and each tool is associated with an action, the action number
directly determines the number of possible plans. Given an action set
$\mathcal{A}$, where $|\mathcal{A}| = n$, the number of possible plans is
$P(n,n)=n!$ if the tested agent strictly follows the user query and invokes
every tool only once. Furthermore, the number of actions also affects the
constraint set size and the generated user query length, as more actions
require more sentences to mention them, which in turn increases the number
of constraints. Both the user query length and the constraint set size
contribute to the difficulty of the problem.

In this RQ, we randomly sample
the value of $|\mathcal{A}|$ from the range of 3 to 5, aligning with the
block number setting in the Blocksworld problem used in the previous
study~\cite{valmeekam2024planning, valmeekam2024planbench}. Blocksworld is
a classic planning problem type adopted by the International Planning
Competition (IPC)~\cite{IPC}, where the agent is required to move stacking
blocks from one configuration to another. Although the difficulty level of
problems in Blocksworld may not directly correspond to the difficulty level
of the problem generated by \tool, even if they share the same difficulty
setting (i.e., the block number equals the action number), we presume that
$|\mathcal{A}|\in [3,5]$ is appropriate for this experiment. Under this
setting, the sub-sentence number of the synthesized user query varies from
one to seven, with the constraint set size ranging from two to ten,
consistent with the common scenario in real-world applications.

\begin{table}[!htpb]
    \centering
    \scriptsize
    \caption{Evaluation results of \tool\ in detecting erroneous planning
        in LLM agents. \textit{Z3-Count} denotes the number of calls to the Z3
        solver. \textit{Time} denotes the total time spent on Z3/synthesis/agent.}
    \label{tab:rq1}
    \setlength{\tabcolsep}{3pt}
    \resizebox{1.0\linewidth}{!}{
        \begin{tabular}{c|l||c|c|c|c|c|c}
            \hline
                                     & \textbf{Agent} & \textbf{\textit{Generated}} & \textbf{\textit{Errors (\% of total)}} & \textbf{\textit{Z3-Count}} & \textbf{\textit{Z3-Time}} & \textbf{\textit{Synthesis-Time}} & \textbf{\textit{Agent-Time}} \\\hline
            \multirow{3}{*}{GPT-3.5} & ReAct          & 843                         & 519 (61.57\%)                          & 99,972                      & 00:24.96                  & 00:59.13                         & 58:00.86                     \\
                                     & OT             & 1,168                        & 635 (54.37\%)                          & 133,008                     & 00:34.30                  & 01:20.49                         & 57:11.94                     \\
                                     & OA             & 327                         & 179 (54.74\%)                          & 36,694                      & 00:09.67                  & 00:22.52                         & 59:22.24                     \\\hline
            \multirow{3}{*}{GPT-4}   & ReAct          & 160                         & 47 (29.38\%)                           & 19,726                      & 00:05.03                  & 00:11.73                         & 59:54.44                     \\
                                     & OT             & 144                         & 32 (22.22\%)                           & 16,253                      & 00:04.13                  & 00:09.66                         & 59:40.41                     \\
                                     & OA             & 111                         & 40 (36.04\%)                           & 12,775                      & 00:03.38                  & 00:07.75                         & 59:54.25                     \\\hline
        \end{tabular}
    }
\end{table}

\T~\ref{tab:rq1} shows the result of the evaluation. The call count and
total time consumed by Z3 are presented in this table to offer a
comprehensive understanding of the overhead associated with
constraint-solving during the synthesis procedure. Moreover, the time
consumption of both the synthesis and agent execution process is
presented. An obvious observation from the table is that \tool\ is highly
effective in synthesizing test cases. The average time consumed to
synthesize a test case is only around 0.07 seconds. Despite the high call
count of Z3, the time spent on Z3 is relatively low, with the average time
spent on Z3 falling below 0.03 seconds per test case. Additionally, less
than half of the total time required for the synthesis process is devoted
to Z3. In contrast, the execution of the agent is the most
time-consuming part, as each agent type takes over 57 minutes.

In general, the speed bottleneck of the entire testing process lies in the agent's
throughput. Both model and framework have a significant impact on the
performance of an LLM agent. Among all agents based on the GPT-3.5 model, ReAct
and OT exhibit a substantial speed advantage over OA, with the number of
processed queries being 843 and 1,168, respectively, compared to OA's 327. For
agents based on the GPT-4 model, the number of processed queries substantially
decreases across all three frameworks. Meanwhile, the gap between different
frameworks narrows, indicating that the token process rate of GPT-4 constitutes
a more severe bottleneck. However, it is important to note that the time
overhead of the agent execution process is negligible in a real-world
scenario, especially for ordinary users. Even in the worst case (OA based on
GPT-4), the agent achieves an average speed of 32.4 seconds per test case. This time
would be negligible given the considerably longer execution times that
real-world tools may require.

In terms of error detection, \tool\ uncovers a considerable number of errors
across all agent settings. Further characterization of the errors will be discussed in RQ2.
From this table, it is evident that GPT-4 significantly improves the agent's performance,
with the error rate of all agents based on GPT-4 decreasing by 48.53\% on average compared to those based on GPT-3.5.
For GPT-3.5, we attribute the comparatively high error rate to its
incapacity to manage the intricate nature of the planning problem, which will be
further investigated in RQ2 and RQ3. Regarding the agent framework, OT is the
most recommended one, outperforming the other two frameworks in both models.

Overall, given that the result check process is carried out in accordance with
the constraint check (see \S~\ref{subsec:verify}), it is guaranteed that any
planning error made by the agent will be detected. Other kinds of errors, such
as invoking a non-existing tool or forgetting to accomplish a task specified in
the query, can also be easily detected by inspecting the execution chain of the
agent.
Given the high comprehensiveness of our testing approach, it is reasonable to
conclude that the error rate reported in \T~\ref{tab:rq1} faithfully reflects
LLM agents' performance. Indeed, the performance of each LLM agent is consistent
with that of previous studies~\cite{yao2022react,shinn2024reflexion}.



\subsection{RQ2: Understanding LLM Agents' Planning Performance and Their Errors}
\label{sec:rq2}


This RQ delves into the planning and errors made by LLM agents.
Recall in RQ1, we explain that LLM agents are generally difficulty-sensitive
systems, and their planning performance would therefore depend on the difficulty
of the encountered problem. Therefore, we presume that only the errors that \textit{fall
within} the planning capability of the LLM agent are valuable for further
investigation. Hence, a comprehensive test with varying difficulties (up to the
upper-level planning capability) is required to exhaustively explore the
planning capability of LLM agents.

\parh{Planning Capability Measurement.}~Specifically, we employ \tool\ to
synthesize extensive test cases, with the difficulty level (i.e., the action number)
gradually increasing. For each difficulty level, we record the success
rate of the planning. We consider the agent to have reached its ``upper-level
planning capability'' when the success rate drops too low (less than 20\%
in our experiments).
Since the sampling space of the query generation is determined by the action
number, we dynamically adjust the sampling number to fit it. That said, for a
given action number $|\mathcal{A}| = n$, we randomly generate $k \times {n
        \choose x}$ test cases, where $k$ is a constant factor and set to 20 in our
experiments. ${n \choose x}$ represents the number of sentences in the extreme case
where the synthesized user query contains sentences that solely mention two actions
in a single sub-sentence (i.e., the possible minimum number of
actions in a sentence). This represents the maximum number of
sentences that can be generated for a given action number, depicting the
sampling space of the query generation.

In our experiments, $n$ starts from 2 and increases by 1 until the success rate
threshold is reached. Likewise, we limit the maximum sampling number for each
action setting to 300, considering the high cost of using OpenAI's API. In this
RQ, we test and present the planning performance of all agents under action numbers
ranging from 2 to 9 to facilitate a comprehensive comparison. Hence, we
conduct the above process for each of the six LLM agents (two models for each of
the three agent frameworks), with 1,600 test cases generated for each agent
type.


\begin{figure}[!htp]
    \centering
    \includegraphics[width=1.0\columnwidth]{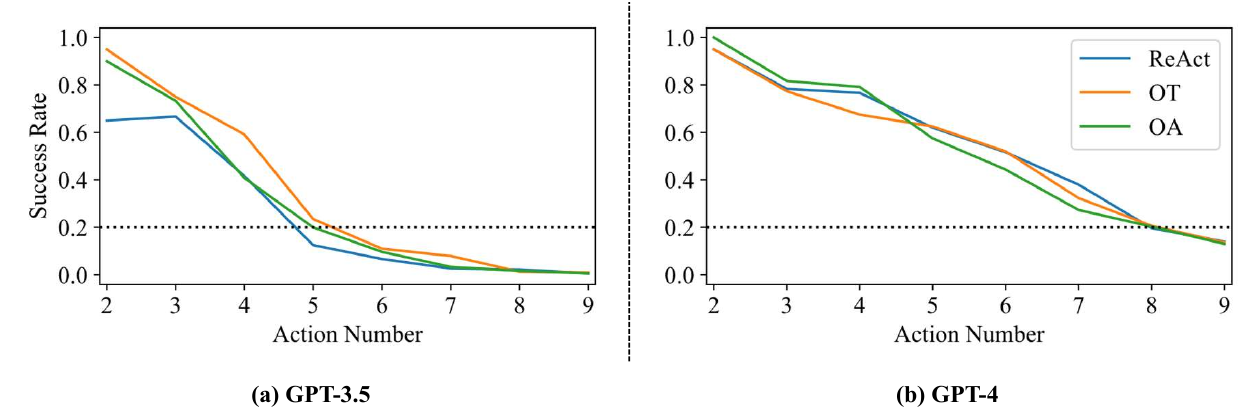}
    \caption{Planning performance of different LLM agents.}
    \label{fig:rq2-1}
\end{figure}

\F~\ref{fig:rq2-1}(a) and \F~\ref{fig:rq2-1}(b) present the planning
success rate of LLM agents based on the GPT-3.5 and GPT-4 models,
respectively. At the 20\% success rate, a dashed line is drawn to signify
the planning capability limit for each agent. Aligning with the findings of
RQ1, agents based on GPT-4 generally exhibit superior planning abilities compared to
those based on GPT-3.5 across all agent frameworks. The upper bound of
agents adopting GPT-4 is achieved when $|\mathcal{A} = 8|$, whereas agents
employing GPT-3.5 reach their maximum planning capability limit when
$|\mathcal{A} = 4|$. Moreover, the success rate of planning for agents based
on GPT-3.5 drops sharply when the action number exceeds three, whereas it
continues to decline gradually for agents based on GPT-4. This suggests
that GPT-4 is more robust in handling complex planning problems. In terms
of the agent framework, it is hard to identify the optimal frameworks in
this setting, as each one has its own strengths. Roughly speaking,
users are recommended to choose OT plus GPT-3.5 for simple tasks and ReAct
plus GPT-4 for complex tasks, considering the time overhead revealed in
RQ1.

\begin{table}[!htpb]
    \centering
    \scriptsize
    \caption{The distribution of different types of planning errors in LLM agents.}
    \label{tab:rq2-1}
    \setlength{\tabcolsep}{3pt}
    \resizebox{0.62\linewidth}{!}{
        \begin{tabular}{c|l||c|c|c|c}
            \hline
                                     & \textbf{Agent} & \textbf{\textit{Timeout}} & \textbf{\textit{Act Error}} & \textbf{\textit{Action Lost}} & \textbf{\textit{Order Error}} \\\hline
            \multirow{3}{*}{GPT-3.5} & ReAct          & 0.00\%                    & 1.03\%                      & 8.25\%                        & 90.72\%                       \\
                                     & OT             & 0.00\%                    & 0.00\%                      & 0.46\%                        & 99.54\%                       \\
                                     & OA             & 0.00\%                    & 0.80\%                      & 0.40\%                        & 98.80\%                       \\\hline
            \multirow{3}{*}{GPT-4}   & ReAct          & 0.00\%                    & 1.27\%                      & 0.00\%                        & 98.73\%                       \\
                                     & OT             & 0.00\%                    & 1.12\%                      & 0.00\%                        & 98.88\%                       \\
                                     & OA             & 0.00\%                    & 1.61\%                      & 0.13\%                        & 98.26\%                       \\\hline
        \end{tabular}
    }
\end{table}

\parh{Error Characteristics.}~We categorize the errors that fall within the
planning capability limit of LLM agents into four types: (1) \textit{Timeout},
where the agent spends too much time (over 180 seconds) or exceeds the maximal
iteration limit (50) without reaching a solution; (2) \textit{Act Error}, where
the agent fails to correctly invoke tools; (3) \textit{Action Lost}, where the
agent fails to accomplish all tasks specified in the query, i.e., some actions
are lost; (4) \textit{Order Error}, where the agent fails to follow the
constraints derived from the query. \T~\ref{tab:rq2-1} presents the error type
distribution for each agent. \textit{Order Error} is the most common error type
across all agents, indicating that LLM agents may encounter difficulties in
untangling the complex constraints of the planning problems. This observation
further confirms the difficulty-sensitive nature of LLM agents. In addition,
ReAct based on GPT-3.5 is substantially more prone to \textit{Action Lost} than
other agents. We attribute this to the lengthy prompt template of ReAct, which
could potentially cause the LLM model to forget some actions specified in the
query. On GPT-4, in contrast, the \textit{Action Lost} error is significantly
reduced, benefiting from the substantial improvement in GPT-4's long-term
memory. We believe these findings are valuable for the design and optimization
of LLM agents (e.g., their accompanying prompt templates), as well as for users
who are interested in adopting LLM agents in their applications.

\begin{table}[!htpb]
    \centering
    \scriptsize
    \caption{The distribution of different root causes in erroneous planning.}
    \label{tab:rq2-2}
    \setlength{\tabcolsep}{3pt}
    \resizebox{0.68\linewidth}{!}{
        \begin{tabular}{c|l||c|c|c|c|c}
            \hline
                                     & \textbf{Agent} & \textbf{\textit{Probability}} & \textbf{\textit{Terminal}} & \textbf{\textit{Topic}} & \textbf{\textit{Structure}} & \textbf{\textit{Constraint}} \\\hline
            \multirow{3}{*}{GPT-3.5} & ReAct          & 19.0\%                        & 50.0\%                     & 18.0\%                  & 9.0\%                       & 4.0\%                        \\
                                     & OT             & 8.0\%                         & 42.0\%                     & 22.0\%                  & 15.0\%                      & 13.0\%                       \\
                                     & OA             & 27.0\%                        & 28.0\%                     & 15.0\%                  & 18.0\%                      & 12.0\%                       \\\hline
            \multirow{3}{*}{GPT-4}   & ReAct          & 19.0\%                        & 44.0\%                     & 14.0\%                  & 18.0\%                      & 5.0\%                        \\
                                     & OT             & 15.0\%                        & 37.0\%                     & 23.0\%                  & 17.0\%                      & 8.0\%                        \\
                                     & OA             & 28.0\%                        & 30.0\%                     & 14.0\%                  & 17.0\%                      & 11.0\%                       \\\hline
        \end{tabular}
    }
\end{table}

\parh{Root Cause Analysis.}~After identifying the planning capability
limit, further investigation into the cause of the triggered failures is
conducted. A systematic error dissection is performed using the
algorithm proposed in \S~\ref{subsec:dissection}. Specifically, we randomly sample
100 errors that fall \textit{within} the planning capability limit and
dissect them to identify the root cause. The error dissection results are
presented in \T~\ref{tab:rq2-2}. Although the root cause distribution
varies across different agent frameworks, it remains notably consistent
across models. The high similarity in error patterns observed among agents based
on different models indicates that agent frameworks
play a more significant role in determining the error cause than the model itself.
Another finding is that \textit{Probability} results in notably more errors in
OA than in other frameworks. We attribute this to the fact that OA cannot
set the temperature of the core LLM, incurring more instability in the
agent's behaviors. Other frameworks, however, still fail to suppress
\textit{Probability} errors to a satisfactory level, even though the sampling
parameters of the core LLM have been tuned before using \tool\ to mitigate non-determinism.
This observation supports our hypothesis discussed in
\S~\ref{subsec:agent-example} that the multi-turn interaction mode of LLM
agents may further exacerbate the non-determinism issue of LLMs. Besides,
it is worth noting that \textit{Terminal} is the most common root cause
across all agents, while \textit{Constraint} typically causes only a small
portion of errors. In other words, altering the prompt through text
mutation not only works well in tweaking LLM's behavior~\cite{jiao2023chatgpt} but can also be
applied to improve or worsen LLM agents' performance.

\begin{table}[!htpb]
    \centering
    \scriptsize
    \caption{Top-5 topics that cause the most errors.}
    \label{tab:rq2-3}
    \setlength{\tabcolsep}{3pt}
    \resizebox{0.65\linewidth}{!}{
        \begin{tabular}{l|c|l|c}
            \hline
            \multicolumn{2}{c|}{\textbf{GPT-3.5}} & \multicolumn{2}{c}{\textbf{GPT-4}}                                   \\\hline
            Topic                                 & count                                & Topic                   & count \\\hline
            Waiter/Waitress                       & 6                                    & Graphic Designer        & 5     \\
            Computer Programmer                   & 5                                    & Electrician             & 4     \\
            Physical Therapist                    & 4                                    & Human Resources Manager & 4     \\
            Marketing Manager                     & 4                                    & Journalist              & 4     \\
            Stock Broker                          & 4                                    & Video Game Designer     & 4     \\\hline
        \end{tabular}
    }
\end{table}

As a further step, since the \textit{Topic} causes a notable number of errors,
we also count the top-5 topics that cause the most errors and present them in
\T~\ref{tab:rq2-3}. Here, we merge the results of agents based on the same
model, as the topic of user queries may solely affect the language comprehension
of the core LLM model. As stated in \S~\ref{subsec:synthesis}, \tool\ randomly
picks one of 50 topics, which are about the daily tasks of different
professions, to fill the action slots in the synthesized user query skeleton. It
is observed that the most challenging topic for GPT3.5 is
\textit{Waiter/Waitress}, which causes 6 errors, while the remaining 19 topics
never cause any errors. Similarly, for GPT-4, \textit{Graphic Designer}'s daily
life seems to be most challenging for the agent (causing 5 errors), while there
are 16 error-free topics. We interpret that the topic of the user query may be
likely unfamiliar to the agent, thereby notably affecting the agent's
performance. Yet, from the perspective of agent users, the role-playing
instruction (currently employed by these agents; introduced in
\S~\ref{subsec:agent-arch}) seems insufficient to improve their performance.

\subsection{RQ3: Further Examining LLM Agents' Performance on Complex Planning Problems}
\label{sec:rq3}


\begin{figure}[htp]
    \centering
    \includegraphics[width=1.0\columnwidth]{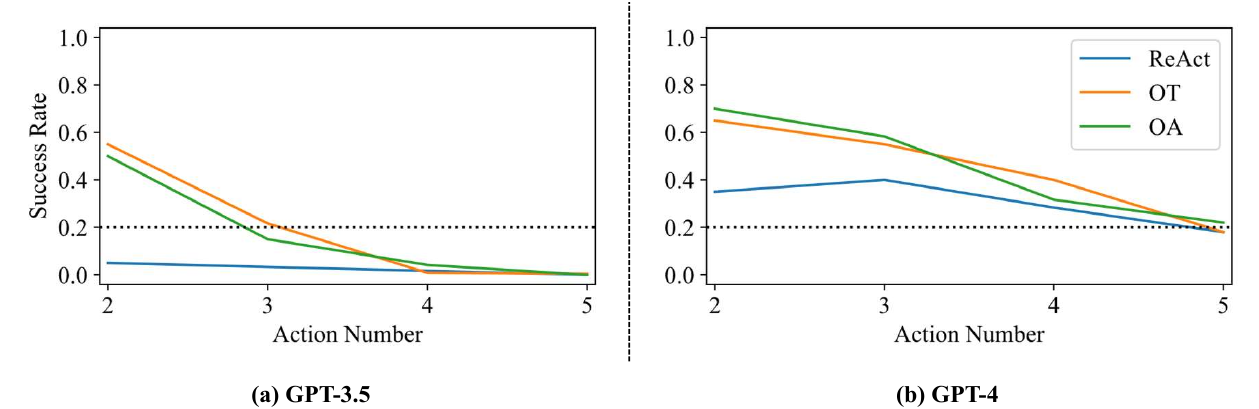}
    \caption{Planning performance of different LLM agents on complex planning problems.}
    \label{fig:rq3-1}
\end{figure}

In this RQ, we repeat the evaluation process described in RQ2 using the extended
version of \tool\ to further examine the planning ability of LLM agents. As stated
in \S~\ref{subsec:extended-version}, the extended version introduces a more
complex planning paradigm by adding \textit{time constraints},
\textit{task duration}, and \textit{tool parameters}, thereby stressing the tested
agents' planning capabilities.

\parh{Planning Capability Measurement.}~We follow the same procedure as
described in RQ2 to measure the upper bound of the LLM agents' planning
capability. \F~\ref{fig:rq3-1} presents the planning success rate of LLM
agents based on the GPT-3.5 and GPT-4 models in this experiment. Since the
maximal planning capability limit of all agents is reached when
$|\mathcal{A} = 5|$, only the planning success rates for action numbers
ranging from 2 to 5 are presented in this figure. Clearly, the success
rate of all agents drops significantly in comparison to the previous
experiment, indicating that the extended version of \tool\ indeed
introduces a more challenging (yet still realistic) planning problem. All agents based on GPT-3.5
drops below the 20\% success rate when the action number reaches three,
suggesting GPT-3.5 may not be suitable for handling such an
intricate planning problem. For agents based on GPT-4, they
maintain a relatively high success rate, though the success rate
suffers a sharp decline when the action number exceeds four. Regarding the
agent framework, ReAct performs the poorest in this setting, failing to
address even the simplest planning problem when it takes GPT-3.5 as the
core. Such failure extends when GPT-4 is adopted, with the success rate
remaining below 50\% across all action numbers. In contrast, OT and OA exhibit
a relatively, albeit limited, higher success rate. We attribute this to the
fact that ReAct relies on few-shot prompting to instruct the model to
invoke tools, and the unstable nature of the LLM model becomes more apparent
when the task becomes more complex. Conversely, OT and OA adopt function
calling, which is implemented by fine-tuning the model with a large number
of function calling data, making them more robust. 

\begin{table}[!htpb]
    \centering
    \scriptsize
    \caption{The distribution of different types of planning errors detected by the extended version of \tool.}
    \label{tab:rq3-1}
    \setlength{\tabcolsep}{3pt}
    \resizebox{0.78\linewidth}{!}{
        \begin{tabular}{c|l||c|c|c|c|c}
            \hline
                                     & \textbf{Agent} & \textbf{\textit{Timeout}} & \textbf{\textit{Act Error}} & \textbf{\textit{Action Lost}} & \textbf{\textit{Order Error}} & \textbf{\textit{Parameter Error}} \\\hline
            \multirow{3}{*}{GPT-3.5} & ReAct          & NaN                       & NaN                         & NaN                           & NaN                           & NaN                               \\
                                     & OT             & 0.0\%                     & 7.14\%                      & 0.0\%                         & 28.57\%                       & 64.29\%                           \\
                                     & OA             & 0.0\%                     & 0.0\%                       & 0.0\%                         & 50.0\%                        & 50.0\%                            \\\hline
            \multirow{3}{*}{GPT-4}   & ReAct          & 0.0\%                     & 64.44\%                     & 0.0\%                         & 33.33\%                       & 2.22\%                            \\
                                     & OT             & 0.0\%                     & 3.77\%                      & 2.83\%                        & 66.04\%                       & 27.36\%                           \\
                                     & OA             & 0.0\%                     & 0.74\%                      & 1.12\%                        & 64.31\%                       & 33.83\%                           \\\hline
        \end{tabular}
    }
\end{table}

\parh{Error Type Analysis.}~\T~\ref{tab:rq3-1} presents the error type
distribution for each agent on the extended version of \tool. The extended
version requires the agent to pass the start time point of actions to the
corresponding tools and understand the tools' return values, which reveal
their execution time. Therefore, we introduce a new error type,
\textit{Parameter Error}, to denote the errors caused by the agent
incorrectly setting the parameters of the tools. In particular, the start
time of one tool should be later than the end time of the previous tool;
otherwise, the agent would be considered to be mismanaging the global
resource --- time --- of the planning problem. From the
table\footnote{ReAct based on GPT-3.5 fails to address the simplest
planning problem in this setting, leading to a NaN error rate in this
table.}, we observe that ReAct encounters a substantial number of
\textit{Act Error}, confirming our earlier interpretation
that ReAct's tool invocation mechanism is not suitable for handling complex
planning problems. For OT and OA, the main error type is \textit{Parameter
Error} when they adopt GPT-3.5, partially explaining their poor performance
in this setting. In contrast, when GPT-4 is adopted, the main error
type for all agents remains \textit{Order Error}, aligning with the
findings in RQ2. In sum, we show that the extended version of \tool\ indeed
introduces a more challenging planning problem, and the agent's performance
is significantly affected by the model and framework. In practice, we
recommend users to employ OT or OA plus GPT-4 to handle such complex
planning problems.

\begin{table}[!htpb]
    \centering
    \scriptsize
    \caption{The distribution of different root causes identified by the extended version of \tool.}
    \label{tab:rq3-2}
    \setlength{\tabcolsep}{3pt}
    \resizebox{0.58\linewidth}{!}{
        \begin{tabular}{l||c|c|c|c|c}
            \hline
                  & \textbf{\textit{Probability}} & \textbf{\textit{Terminal}} & \textbf{\textit{Topic}} & \textbf{\textit{Structure}} & \textbf{\textit{Constraint}} \\\hline
            ReAct & 8.0\%                         & 22.0\%                     & 11.0\%                  & 14.0\%                      & 45.0\%                       \\
            OT    & 25.0\%                        & 32.0\%                     & 12.0\%                  & 8.0\%                       & 23.0\%                       \\
            OA    & 23.0\%                        & 22.0\%                     & 11.0\%                  & 5.0\%                       & 39.0\%                       \\\hline
        \end{tabular}
    }
\end{table}

\begin{figure}[!htp]
    \centering
    \includegraphics[width=0.75\columnwidth]{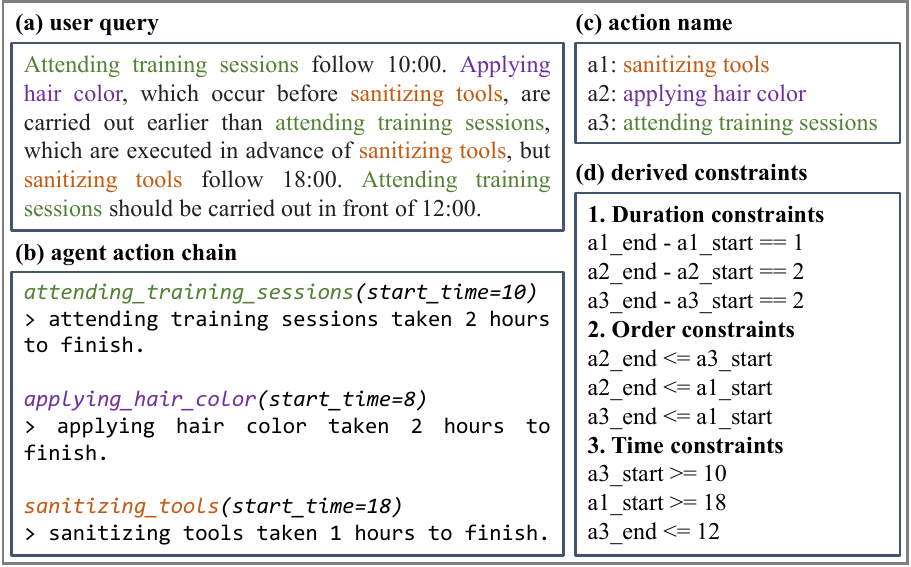}
    \caption{An example of the erroneous planning made by an LLM agent.
        Actions are highlighted in different colors. To facilitate the
        understanding, we also present the derived constraints and the mapping
        between the actions and the variables in the constraints.}
    \label{fig:rq3-2}
\end{figure}

\parh{Root Cause Analysis.}~Similar to RQ2, we dissect the errors made by
LLM agents. Due to the failure of GPT-3.5-based agents on the extended
version of \tool, we primarily inspect the root cause for GPT-4-based agents 
to provide a more insightful analysis.
\T~\ref{tab:rq3-2} presents the root cause distribution for each agent. A
notable difference from the results in RQ2 is that \textit{Constraint}
causes a larger portion of errors in this experiment, taking the dominant
position in ReAct and OA, and ranking second in OT. We believe this is
due to several reasons as follows.

First, the agent is required to adjust its plan dynamically according to the
tool return values, which substantially increases the complexity of the planning
problem as discussed in \S~\ref{subsec:extended-version}. Second, the experiment
results reveal that LLMs tend to make more errors when handling multiple
different kinds of requirements simultaneously, which is a common scenario in
real-world applications. \F~\ref{fig:rq3-2} provides an example of the erroneous
planning made by an LLM agent. In this example, the agent fails to pass the
correct start time to \texttt{applying_hair_color}, leading to a
\textit{Parameter Error}. Since the previous tool started at 10:00 and took two
hours, \texttt{applying_hair_color} should start after 12:00 (i.e., $start\_time
\geq 12$), but the agent mistakenly sets the start time to 8. Although the
action chain would be correct if the agent could reorder the actions according
to the current start time of the tools, it already violates the requirement that
``You need to execute all tasks one by one in the correct order and within the
time range set by the requirement'' in the query (see
\F~\ref{fig:extended-exp}). Furthermore, the introduction of time and duration
undoubtedly adds more constraints to the planning problem, which may also
exacerbate agents' planning difficulties.

\section{Discussion}
\label{sec:discussion}

\parh{Alternative Approaches for Test Case Generation.}~Several other approaches
can be employed to generate test cases for LLM agents, including text mutation
and text generation based on LLM. Based on preliminary study and experiments, we
find that these methods are much less effective than our approach. In short,
while these methods may seem simpler to implement and lighter in weight, they
are incapable of generating high-diversity test cases and guaranteeing the
correctness of result verification in the meantime. 

Text mutation is limited to superficial changes in the input text, which is
hardly feasible to cover the diverse scenarios that LLM agents may encounter. 
As for text generation, a possible approach is feeding a constraint set into an
LLM, and ask the LLM to generate a user query in natural language. This method,
nevertheless, is problematic because of the unstable nature of LLMs. To
illustrate, we carry out an experiment in which we feed a constraint set and the
variable-action mapping (e.g., $\cf{a_1}$: \texttt{network\_diagnose}) into
GPT-4 and instruct it to generate the corresponding user query. Then, the
generated user query is manually verified to see if it is semantically
consistent with the given constraints. For action numbers ranging from 2 to 5,
we repeat the experiment 10 times for each action number. The results show that
the quality of the generation is negatively affected by the action number
$|\mathcal{A}|$, which reflects the complexity of the planning problem. In
particular, the generation attains a 100\% accuracy rate when the action number
is below four, whereas it falls to 80\% when $|\mathcal{A} =4|$ and to 0\% when
$|\mathcal{A} =5|$. Considering that the test effectiveness is determined by the
consistency of the generated text with the constraints, the results suggest that
the text generation approach is unsuitable for this task.

\parh{Threat to Validity.}~Our approach is based on the assumption that no
inherent sequential dependencies exist among the actions in the planning
tasks, i.e., the derived constraints wholly reflect the constraints among
the actions. This assumption shall hold for most cases, but there exist
scenarios where the inherent sequential dependencies contained by the
actions are common sense and do not need to be explicitly specified. In
such cases, our approach may fail if there is a conflict between the
inherent dependencies and the constraints derived from the synthesized
plans. We leave it as future work to investigate how to handle such corner
cases. 


\parh{Extensions.}~Our approach can be extended in several directions. First, we
can explore the use of other LLMs like Claude-3~\cite{claude3} and other agent
frameworks. Given the rather limited performance of current state-of-the-art
LLMs when being tested by the extended version of \tool, we presume it is
necessary to further gauge the planning capabilities of further advanced LLMs.
Integrating \tool\ with other mainstream agent frameworks and LLMs is
straightforward, as \tool\ does not rely on specific LLM or agent framework
features.

Another straightforward extension is to apply our approach to fine-tune the LLMs
for planning tasks. The intuition is that \tool\ can generate numerous diverse
and high-quality test cases with different complexities. More importantly,
\tool\ can automatically identify which constraints are violated by the LLMs,
and therefore, it provides valuable feedback to the LLMs to improve their
planning capabilities. Such an ``error-driven'' fine-tuning process can be
repeated iteratively to enhance the LLMs' planning capabilities. knowledgeable
audience may be aware that localizing which constraints are violated by the LLMs
can be smoothly implemented by using the \texttt{unsat\_core} function in the Z3
solver.

One may also question the feasibility of repairing the detected erroneous plans
in an ``on-the-fly'' manner, where we use constraint solvers to generate a new
plan that satisfies the requirements. This can be achieved by first extracting
the constraints from an error-triggering user query encountered during daily
agent usage (not those test query inputs synthesized by \tool), then using Z3 to
find a plan that satisfies the constraints. 
We clarify, however, that extracting the constraints from arbitrary natural
language queries in the wild is still in its infancy, despite several
encouraging progresses~\cite{liu2023llm, ye2024satlm} have been made. An initial
consideration is that the extracted constraints might be inconsistent or
insufficient due to the unstable nature of LLMs. Furthermore, the intricate
interaction mode --- the user query, the tool's name and description, and the
return message, all of which may contain the constraints --- may complicate the
constraint extraction process. In contrast, previous work~\cite{liu2023llm,
ye2024satlm} mainly focuses on testing the LLM, rather than the agent, on
well-formed planning questions, neglecting the real-world scenario where the LLM
agent is required to interact with the user.

\section{Related Work}
\label{sec:related}

\parh{Benchmarking LLM Agents.}~There has been a growing interest in
benchmarking LLM agents, and several benchmarking suites have been
proposed.
Wu et al.~\cite{wu2023smartplay} proposed SmartPlay, a benchmarking suite
that contains multiple specialized games, aiming to evaluate the
understanding, knowledge, and reasoning capabilities of LLM agents in a
comprehensive manner. Trulens~\cite{trulens} is another benchmarking suite
that is designed to evaluate the performance of LLM agents on tasks that
require complex reasoning and planning. Besides, a series of studies have
been conducted to gauge the LLM agent from different perspectives,
including tool interaction~\cite{huang2023metatool}, robustness against
jailbreak~\cite{zhan2024injecagent}, and safety risk
awareness~\cite{naihin2023testing, yuan2024r}.

\parh{Testing LLMs.}~In line with their remarkable success in various
applications, LLMs have been emergingly tested to ensure their reliability and
robustness across different scenarios. A recent trend in testing LLMs is to
gauge their logical reasoning capabilities, including mathematical
reasoning~\cite{stolfo2022causal}, causal inference~\cite{kiciman2023causal,
long2023can}, and planning~\cite{valmeekam2024planbench,valmeekam2024planning}.
Among these, the planning capability is particularly crucial for LLM as it
constitutes the foundation for many applications, such as autonomous
vehicles~\cite{LLMagentLilian23, chen2023interactive},
robotics~\cite{joshi2023locally}, and any agent-based
systems~\cite{yao2022react, shao2023character, wang2024describe}.



\section{Conclusion}

We have proposed \tool, a novel and automated approach to testing and
understanding the planning ability of LLM agents. We formulate the
detection of erroneous planning as a constraint satisfiability problem and
propose a fully automated framework to synthesize diverse and high-quality
user inputs for testing. We evaluate \tool's effectiveness using three
mainstream agent frameworks and two powerful LLMs (GPT-3.5 and GPT-4).
The results show that \tool\ can effectively detect LLM agents' erroneous
planning and provide valuable insights into the mechanisms underlying the
errors.

\bibliographystyle{acm}
\bibliography{bib/causality,bib/machine-learning,bib/ref,bib/llm,bib/pl,bib/zj}

\begin{thebibliography}{10}

\bibitem{trulens}
Trulens, evaluating and testing llm apps.
\newblock \url{https://medium.com/trulens}.

\bibitem{agentfinancial}
Designing llm agent tools for due diligence in financial instruments.
\newblock \url{https://developers.lseg.com/en/article-catalog/article/designing-llm-agent-tools-for-due-diligence-in-financial-instruments}, 2024.

\bibitem{IPC}
International planning competition.
\newblock \url{https://www.icaps-conference.org/competitions/}, 2024.

\bibitem{claude3}
Introducing the next generation of claude.
\newblock \url{https://www.anthropic.com/news/claude-3-family}, 2024.

\bibitem{openaiTools}
Openai tools.
\newblock \url{https://python.langchain.com/docs/modules/agents/agent_types/openai_tools}, 2024.

\bibitem{openaiFC}
Openai's function callling.
\newblock \url{https://platform.openai.com/docs/guides/function-calling}, 2024.

\bibitem{openaiAssistant}
Overview of openai's assistant.
\newblock \url{https://platform.openai.com/docs/assistants/overview?context=with-streaming}, 2024.

\bibitem{agentprove}
This ai research introduces ‘rafa’: A principled artificial intelligence framework for autonomous llm agents with provable sample efficiency.
\newblock \url{https://www.marktechpost.com/2023/10/24/this-ai-research-introduces-rafa-a-principled-artificial-intelligence-framework-for-autonomous-llm-agents-with-provable-sample-efficiency/}, 2024.

\bibitem{unskript}
unskript launches ai-powered infrastructure health intelligence platform for software teams.
\newblock \url{https://markets.businessinsider.com/news/stocks/unskript-launches-ai-powered-infrastructure-health-intelligence-platform-for-software-teams-1032992108}, 2024.

\bibitem{achiam2023gpt}
{\sc Achiam, J., Adler, S., Agarwal, S., Ahmad, L., Akkaya, I., Aleman, F.~L., Almeida, D., Altenschmidt, J., Altman, S., Anadkat, S., et~al.}
\newblock Gpt-4 technical report.
\newblock {\em arXiv preprint arXiv:2303.08774\/} (2023).

\bibitem{ahn2022can}
{\sc Ahn, M., Brohan, A., Brown, N., Chebotar, Y., Cortes, O., David, B., Finn, C., Fu, C., Gopalakrishnan, K., Hausman, K., et~al.}
\newblock Do as i can, not as i say: Grounding language in robotic affordances.
\newblock {\em arXiv preprint arXiv:2204.01691\/} (2022).

\bibitem{bills2023language}
{\sc Bills, S., Cammarata, N., Mossing, D., Tillman, H., Gao, L., Goh, G., Sutskever, I., Leike, J., Wu, J., and Saunders, W.}
\newblock Language models can explain neurons in language models.
\newblock {\em \url{https://openaipublic.blob.core.windows.net/neuron-explainer/paper/index.html}\/} (2023).

\bibitem{carbonell1991prodigy}
{\sc Carbonell, J., Etzioni, O., Gil, Y., Joseph, R., Knoblock, C., Minton, S., and Veloso, M.}
\newblock Prodigy: An integrated architecture for planning and learning.
\newblock {\em ACM SIGART Bulletin 2}, 4 (1991), 51--55.

\bibitem{LangChain}
{\sc Chase, H.}
\newblock {LangChain}, Oct. 2022.

\bibitem{chen2023interactive}
{\sc Chen, Y., Veer, S., Karkus, P., and Pavone, M.}
\newblock Interactive joint planning for autonomous vehicles.
\newblock {\em IEEE Robotics and Automation Letters\/} (2023).

\bibitem{creswell2022selection}
{\sc Creswell, A., Shanahan, M., and Higgins, I.}
\newblock Selection-inference: Exploiting large language models for interpretable logical reasoning.
\newblock {\em arXiv preprint arXiv:2205.09712\/} (2022).

\bibitem{de2008z3}
{\sc De~Moura, L., and Bj{\o}rner, N.}
\newblock Z3: An efficient smt solver.
\newblock In {\em International conference on Tools and Algorithms for the Construction and Analysis of Systems\/} (2008), Springer, pp.~337--340.

\bibitem{ghallab2004automated}
{\sc Ghallab, M., Nau, D., and Traverso, P.}
\newblock {\em Automated Planning: theory and practice}.
\newblock Elsevier, 2004.

\bibitem{CannotSelfCorrect24}
{\sc Huang, J., Chen, X., Mishra, S., Zheng, H.~S., Yu, A.~W., Song, X., and Zhou, D.}
\newblock Large language models cannot self-correct reasoning yet.
\newblock In {\em Proc. ICLR\/} (2024).

\bibitem{huang2023metatool}
{\sc Huang, Y., Shi, J., Li, Y., Fan, C., Wu, S., Zhang, Q., Liu, Y., Zhou, P., Wan, Y., Gong, N.~Z., et~al.}
\newblock Metatool benchmark for large language models: Deciding whether to use tools and which to use.
\newblock {\em arXiv preprint arXiv:2310.03128\/} (2023).

\bibitem{CausalityLLMCode23}
{\sc Ji, Z., Ma, P., Li, Z., and Wang, S.}
\newblock Benchmarking and explaining large language model-based code generation: A causality-centric approach.
\newblock {\em arXiv preprint arXiv:2310.06680\/} (2023).

\bibitem{jiao2023chatgpt}
{\sc Jiao, W., Wang, W., Huang, J.-t., Wang, X., Shi, S., and Tu, Z.}
\newblock {Is ChatGPT A Good Translator? Yes With GPT-4 As The Engine}.
\newblock {\em arXiv preprint arXiv:2301.08745\/} (2023).

\bibitem{joshi2023locally}
{\sc Joshi, S.~S., Hutchinson, S., and Tsiotras, P.}
\newblock Les: Locally exploitative sampling for robot path planning.
\newblock In {\em 2023 IEEE International Conference on Robotics and Automation (ICRA)\/} (2023), IEEE, pp.~1551--1557.

\bibitem{kiciman2023causal}
{\sc K{\i}c{\i}man, E., Ness, R., Sharma, A., and Tan, C.}
\newblock Causal reasoning and large language models: Opening a new frontier for causality.
\newblock {\em arXiv preprint arXiv:2305.00050\/} (2023).

\bibitem{CoTFaithfulness23}
{\sc Lanham, T., Chen, A., Radhakrishnan, A., Steiner, B., Denison, C., Hernandez, D., Li, D., Durmus, E., Hubinger, E., Kernion, J., Lukosiute, K., Nguyen, K., Cheng, N., Joseph, N., Schiefer, N., Rausch, O., Larson, R., McCandlish, S., Kundu, S., Kadavath, S., Yang, S., Henighan, T., Maxwell, T., Telleen{-}Lawton, T., Hume, T., Hatfield{-}Dodds, Z., Kaplan, J., Brauner, J., Bowman, S.~R., and Perez, E.}
\newblock Measuring faithfulness in chain-of-thought reasoning.
\newblock {\em arXiv preprint arXiv:2307.13702\/} (2023).

\bibitem{li2022cctest}
{\sc Li, Z., Wang, C., Liu, Z., Wang, H., Wang, S., and Gao, C.}
\newblock {CCTEST}: Testing and repairing code completion systems.
\newblock In {\em Proc. ACM ICSE\/} (2023).

\bibitem{li2023split}
{\sc Li, Z., Wang, C., Ma, P., Wu, D., Li, T., Wang, S., Gao, C., and Liu, Y.}
\newblock Split and merge: Aligning position biases in large language model based evaluators.
\newblock {\em arXiv preprint arXiv:2310.01432\/} (2023).

\bibitem{CriticBench24}
{\sc Lin, Z., Gou, Z., Liang, T., Luo, R., Liu, H., and Yang, Y.}
\newblock {CriticBench: Benchmarking LLMs for Critique-Correct Reasoning}.
\newblock {\em arXiv preprint arXiv:2402.14809\/} (2024).

\bibitem{liu2023llm}
{\sc Liu, B., Jiang, Y., Zhang, X., Liu, Q., Zhang, S., Biswas, J., and Stone, P.}
\newblock Llm+ p: Empowering large language models with optimal planning proficiency.
\newblock {\em arXiv preprint arXiv:2304.11477\/} (2023).

\bibitem{liu2021efficient}
{\sc Liu, S., Lu, N., Chen, C., and Tang, K.}
\newblock Efficient combinatorial optimization for word-level adversarial textual attack.
\newblock {\em IEEE/ACM Transactions on Audio, Speech, and Language Processing 30\/} (2021), 98--111.

\bibitem{long2023can}
{\sc Long, S., Schuster, T., Pich{\'e}, A., de~Montreal, U., Research, S., et~al.}
\newblock Can large language models build causal graphs?
\newblock {\em arXiv preprint arXiv:2303.05279\/} (2023).

\bibitem{lu2023learning}
{\sc Lu, Y., Zhang, X., Xu, X., and Yao, W.}
\newblock Learning-based near-optimal motion planning for intelligent vehicles with uncertain dynamics.
\newblock {\em IEEE Robotics and Automation Letters\/} (2023).

\bibitem{iAudit24}
{\sc Ma, W., Wu, D., Sun, Y., Wang, T., Liu, S., Zhang, J., Xue, Y., and Liu, Y.}
\newblock Combining fine-tuning and {LLM-based} agents for intuitive smart contract auditing with justifications.
\newblock {\em arXiv preprint arXiv:2403.16073\/} (2024).

\bibitem{mccarthy1963situations}
{\sc McCarthy, J., et~al.}
\newblock {\em Situations, actions, and causal laws}.
\newblock Comtex Scientific, 1963.

\bibitem{naihin2023testing}
{\sc Naihin, S., Atkinson, D., Green, M., Hamadi, M., Swift, C., Schonholtz, D., Kalai, A.~T., and Bau, D.}
\newblock Testing language model agents safely in the wild.
\newblock {\em arXiv preprint arXiv:2311.10538\/} (2023).

\bibitem{nilsson1984shakey}
{\sc Nilsson, N.~J., et~al.}
\newblock {\em Shakey the robot}, vol.~323.
\newblock Sri International Menlo Park, California, 1984.

\bibitem{papineni2002bleu}
{\sc Papineni, K., Roukos, S., Ward, T., and Zhu, W.-J.}
\newblock Bleu: a method for automatic evaluation of machine translation.
\newblock In {\em Proceedings of the 40th annual meeting of the Association for Computational Linguistics\/} (2002), pp.~311--318.

\bibitem{LLMLogicalFallacies23}
{\sc Payandeh, A., Pluth, D., Hosier, J., Xiao, X., and Gurbani, V.~K.}
\newblock How susceptible are {LLMs} to logical fallacies?
\newblock {\em arXiv preprint arXiv:2308.09853\/} (2023).

\bibitem{shao2023character}
{\sc Shao, Y., Li, L., Dai, J., and Qiu, X.}
\newblock Character-llm: A trainable agent for role-playing.
\newblock {\em arXiv preprint arXiv:2310.10158\/} (2023).

\bibitem{shen2024hugginggpt}
{\sc Shen, Y., Song, K., Tan, X., Li, D., Lu, W., and Zhuang, Y.}
\newblock Hugginggpt: Solving ai tasks with chatgpt and its friends in hugging face.
\newblock {\em Advances in Neural Information Processing Systems 36\/} (2024).

\bibitem{shinn2024reflexion}
{\sc Shinn, N., Cassano, F., Gopinath, A., Narasimhan, K., and Yao, S.}
\newblock Reflexion: Language agents with verbal reinforcement learning.
\newblock {\em Advances in Neural Information Processing Systems 36\/} (2024).

\bibitem{shridhar2020alfworld}
{\sc Shridhar, M., Yuan, X., C{\^o}t{\'e}, M.-A., Bisk, Y., Trischler, A., and Hausknecht, M.}
\newblock Alfworld: Aligning text and embodied environments for interactive learning.
\newblock {\em arXiv preprint arXiv:2010.03768\/} (2020).

\bibitem{AutoGPT}
{\sc {Significant Gravitas}}.
\newblock {AutoGPT}.

\bibitem{singhal2023large}
{\sc Singhal, K., Azizi, S., Tu, T., Mahdavi, S.~S., Wei, J., Chung, H.~W., Scales, N., Tanwani, A., Cole-Lewis, H., Pfohl, S., et~al.}
\newblock Large language models encode clinical knowledge.
\newblock {\em Nature\/} (2023), 1--9.

\bibitem{stolfo2022causal}
{\sc Stolfo, A., Jin, Z., Shridhar, K., Sch{\"o}lkopf, B., and Sachan, M.}
\newblock A causal framework to quantify the robustness of mathematical reasoning with language models.
\newblock {\em arXiv preprint arXiv:2210.12023\/} (2022).

\bibitem{LLM4Vuln24}
{\sc Sun, Y., Wu, D., Xue, Y., Liu, H., Ma, W., Zhang, L., Shi, M., and Liu, Y.}
\newblock {LLM4Vuln}: A unified evaluation framework for decoupling and enhancing llms' vulnerability reasoning.
\newblock {\em arXiv preprint arXiv:2401.16185\/} (2024).

\bibitem{GPTScan24}
{\sc Sun, Y., Wu, D., Xue, Y., Liu, H., Wang, H., Xu, Z., Xie, X., and Liu, Y.}
\newblock {GPTScan}: {Detecting} logic vulnerabilities in smart contracts by combining {GPT} with program analysis.
\newblock In {\em Proc. ACM ICSE\/} (2024).

\bibitem{sun2020automatic}
{\sc Sun, Z., Zhang, J.~M., Harman, M., Papadakis, M., and Zhang, L.}
\newblock Automatic testing and improvement of machine translation.
\newblock In {\em Proceedings of the ACM/IEEE 42nd International Conference on Software Engineering\/} (2020), pp.~974--985.

\bibitem{valmeekam2024planbench}
{\sc Valmeekam, K., Marquez, M., Olmo, A., Sreedharan, S., and Kambhampati, S.}
\newblock {PlanBench}: An extensible benchmark for evaluating large language models on planning and reasoning about change.
\newblock {\em Advances in Neural Information Processing Systems 36\/} (2024).

\bibitem{valmeekam2024planning}
{\sc Valmeekam, K., Marquez, M., Sreedharan, S., and Kambhampati, S.}
\newblock On the planning abilities of large language models-a critical investigation.
\newblock {\em Advances in Neural Information Processing Systems 36\/} (2024).

\bibitem{ReasonWithRules24}
{\sc Wang, S., Wei, Z., Choi, Y., and Ren, X.}
\newblock {Can LLMs Reason with Rules? Logic Scaffolding for Stress-Testing and Improving LLMs}.
\newblock {\em arXiv preprint arXiv:2402.11442\/} (2024).

\bibitem{wang2024describe}
{\sc Wang, Z., Cai, S., Chen, G., Liu, A., Ma, X.~S., and Liang, Y.}
\newblock Describe, explain, plan and select: interactive planning with llms enables open-world multi-task agents.
\newblock {\em Advances in Neural Information Processing Systems 36\/} (2024).

\bibitem{Jailbroken23}
{\sc Wei, A., Haghtalab, N., and Steinhardt, J.}
\newblock Jailbroken: How does {LLM} safety training fail?
\newblock In {\em Proc. NeurIPS\/} (2023).

\bibitem{LLMagentLilian23}
{\sc Weng, L.}
\newblock {LLM Powered Autonomous Agents}.
\newblock \url{https://lilianweng.github.io/posts/2023-06-23-agent/}, 2023.

\bibitem{DeGPT23}
{\sc Wong, W.~K., Wang, H., Li, Z., Liu, Z., Wang, S., Tang, Q., Nie, S., and Wu, S.}
\newblock Refining decompiled c code with large language models.
\newblock {\em arXiv preprint arXiv:2310.06530\/} (2023).

\bibitem{wu2024symbol}
{\sc Wu, X., Li, Y.-L., Sun, J., and Lu, C.}
\newblock Symbol-llm: Leverage language models for symbolic system in visual human activity reasoning.
\newblock {\em Advances in Neural Information Processing Systems 36\/} (2024).

\bibitem{wu2023smartplay}
{\sc Wu, Y., Tang, X., Mitchell, T.~M., and Li, Y.}
\newblock Smartplay: A benchmark for llms as intelligent agents.
\newblock {\em arXiv preprint arXiv:2310.01557\/} (2023).

\bibitem{HallucinationInevitable24}
{\sc Xu, Z., Jain, S., and Kankanhalli, M.}
\newblock Hallucination is inevitable: An innate limitation of large language models.
\newblock {\em arXiv preprint arXiv:2401.11817\/} (2023).

\bibitem{yao2022react}
{\sc Yao, S., Zhao, J., Yu, D., Du, N., Shafran, I., Narasimhan, K., and Cao, Y.}
\newblock React: Synergizing reasoning and acting in language models.
\newblock {\em arXiv preprint arXiv:2210.03629\/} (2022).

\bibitem{ye2024satlm}
{\sc Ye, X., Chen, Q., Dillig, I., and Durrett, G.}
\newblock Satlm: Satisfiability-aided language models using declarative prompting.
\newblock {\em Advances in Neural Information Processing Systems 36\/} (2024).

\bibitem{yuan2024r}
{\sc Yuan, T., He, Z., Dong, L., Wang, Y., Zhao, R., Xia, T., Xu, L., Zhou, B., Li, F., Zhang, Z., et~al.}
\newblock R-judge: Benchmarking safety risk awareness for llm agents.
\newblock {\em arXiv preprint arXiv:2401.10019\/} (2024).

\bibitem{EvaluateInstructionFollowing23}
{\sc Zeng, Z., Yu, J., Gao, T., Meng, Y., Goyal, T., and Chen, D.}
\newblock Evaluating large language models at evaluating instruction following.
\newblock {\em arXiv preprint arXiv:2310.07641\/} (2023).

\bibitem{zhan2024injecagent}
{\sc Zhan, Q., Liang, Z., Ying, Z., and Kang, D.}
\newblock Injecagent: Benchmarking indirect prompt injections in tool-integrated large language model agents.
\newblock {\em arXiv preprint arXiv:2403.02691\/} (2024).

\bibitem{zhang2019bertscore}
{\sc Zhang, T., Kishore, V., Wu, F., Weinberger, K.~Q., and Artzi, Y.}
\newblock Bertscore: Evaluating text generation with bert.
\newblock {\em arXiv preprint arXiv:1904.09675\/} (2019).

\end{thebibliography}

\end{document}